\documentclass[]{article}
\usepackage{proceed2e}

\usepackage{times}
\usepackage{graphicx}
\usepackage{subfigure}
\usepackage[round]{natbib}
\usepackage{amsmath}
\usepackage{amssymb}
\usepackage{amsbsy}
\usepackage{defs}
\usepackage{algorithm}
\usepackage{algorithmic}

\title{Bayesian Structure Learning for Markov Random Fields \\with a Spike and Slab Prior}

\author{ {\bf Yutian Chen} \\  
Department of Computer Science\\  
University of California, Irvine\\ 
Irvine, CA 92697\\ 
\And 
{\bf Max Welling} \\
Department of Computer Science\\  
University of California, Irvine\\ 
Irvine, CA 92697\\
} 
 
\begin{document} 
 
\maketitle 
 
\begin{abstract} 
In recent years a number of methods have been developed for automatically learning the (sparse) connectivity structure of Markov Random Fields. These methods are mostly based on $L_1$-regularized optimization which has a number of disadvantages such as the inability to assess model uncertainty and expensive cross-validation to find the optimal regularization parameter. Moreover, the model's predictive performance may degrade dramatically with a suboptimal value of the regularization parameter (which is sometimes desirable to induce sparseness). We propose a fully Bayesian approach based on a ``spike and slab" prior (similar to $L_0$ regularization) that does not suffer from these shortcomings. We develop an approximate MCMC method combining Langevin dynamics and reversible jump MCMC to conduct inference in this model. Experiments show that the proposed model learns a good combination of the structure and parameter values without the need for separate hyper-parameter tuning. Moreover, the model's predictive performance is much more robust than $L_1$-based methods with hyper-parameter settings that induce highly sparse model structures.
\end{abstract} 

\section{Introduction}
Undirected probabilistic graphical models, also known as Markov Random Fields (MRFs), have been widely used in a large variety of domains including computer vision \citep{li2009markov}, natural language processing \citep{sha2003shallow}, and social networks \citep{robins2007introduction}. The structure of the model is defined through a set of features defined on subsets of random variables. Automated methods to select relevant features are becoming increasingly important in a time where the proliferation of sensors make it possible to measure a multitude of data-attributes. There is also an increasing interest in \emph{sparse} model structures because they help against overfitting and are computationally more tractable than dense model structures.

In this paper we focus on a particular type of MRF, called a log-linear model, where structure learning or feature selection is integrated with parameter estimation. Although structure learning has been extensively studied for directed graphical models, it is typically more difficult for undirected models due to the intractable normalization term of the probability distribution, known as the partition function. Traditional algorithms apply only to restricted types of structures with low tree-width \citep{andrew2007scalable,tsuruoka2009stochastic,hu2009log} or special models such as Gaussian graphical models \citep{jones2005experiments} so that accurate inference can be conducted efficiently.

For an arbitrary structure, various methods have been proposed in the literature, generally categorized into two approaches. One approach is based on separate tests on an edge or the neighbourhood of a node so that there is no need to compute the joint distribution \citep{wainwright2007high,bresler2008reconstruction,ravikumar2010high}. The other approach is based on maximum likelihood estimation (MLE) with a sparsity inducing criterion. These methods require approximate inference algorithms in order to estimate the log-likelihood such as Gibbs sampling \citep{della1997inducing}, loopy belief propagation \citep{lee2006efficient,zhu2010grafting}, or pseudo-likelihood \citep{höfling2009estimation}. A popular choice of such a criterion is $L_1$ regularization \citep{riezler2004incremental,dudik2004performance} which enjoys several good properties such as a convex objective function and a consistency guarantee. However, $L_1$-regularized MLE is usually sensitive to the choice the regularization strength, and these optimization-based methods cannot provide a credible interval for the learned structure. Also, in order to learn a sparse structure, a strong penalty has to be imposed on all the edges which usually results in suboptimal parameter values.

We will follow a third approach to MRF structure learning in a fully Bayesian framework which has not been explored yet. The Bayesian approach considers the structure of a graphical model as random. Inference in a Bayesian model provides inherent regularization, and offers a fully probabilistic characterization of the underlying structure. It was shown in \citet{park2008bayesian} that Bayesian models with a Gaussian or Laplace prior distribution (corresponding to $L_2$ or $L_1$ regularization) do not exhibit sparse structure. \citet{mohamed2011bayesian} proposes to use a ``spike and slab" prior for learning \emph{directed} graphical models which corresponds to the ideal $L_0$ regularization. This model exhibits better robustness against over-fitting than the related $L_1$ approaches. Unlike the Laplace/Gaussian prior, the posterior distribution over parameters for a ``spike and slab" prior is no longer guaranteed to be unimodal. However, approximate inference methods have been successfully applied in the context of directed models using MCMC \citep{mohamed2011bayesian} and expectation propagation \citep{hernández2010expectation}.

Unfortunately, Bayesian inference for MRFs is much harder than for directed networks due to the partition function. This feature renders even MCMC sampling intractable which caused some people to dub these problems ``double intractability" \citep{MurrayGhahramaniMacKay06}. Nevertheless, variational methods \citep{parise2006structure,QiSzummerMinka05} and MCMC methods \citep{MurrayGhahramani04} have been successfully explored for approximate inference when the model structure is fixed.

We propose a Bayesian structure learning method with a spike and slab prior for MRFs and devise an approximate MCMC method to draw samples of both the model structure and the model parameters by combining a modified Langevin sampler with a reversible jump MCMC method. Experiments show that the posterior distribution estimated by our inference method matches the actual distribution very well. Moreover, our method offers better robustness to both under-fitting and over-fitting than $L_1$-regularized methods. A related but different application of the spike and slab distribution in MRFs is shown in \citet{courville2011spike} for modelling hidden random variables.

This paper is organized as follows: we first introduce a hierarchical Bayesian model for MRF structure learning in section \ref{sec:model} and then describe an approximate MCMC method in section \ref{sec:lg}, \ref{sec:rjmcmc}, and \ref{sec:others} to draw samples for the model parameters, structure, and other hyper-parameters respectively. Experiments are conducted in section \ref{sec:experiments} on two simulated data sets and a real-world dataset, followed by a discussion section.

\section{Learning the Structure of MRFs as Bayesian Inference} \label{sec:model}
The probability distribution of a MRF is defined by a set of potential functions. Consider a widely used family of MRFs with log-linear parametrization:
\be
P(\bx|\bta) = \frac{1}{Z(\bta)} \exp\left(\sum_{\al} \ta_{\al} f_{\al}(\bx_{\al})\right) \label{eqn:mrf}
\ee
where each potential function is defined as the exponential of the product between a feature function $f_{\al}$ of a set of variables $\bx_{\al}$ and an associated parameter $\ta_{\al}$. $Z$ is called the partition function. All the variables in the scope of a potential function form a clique in their graphical representation. When a parameter $\ta_{\al}$ has a value of zero, we could equivalently remove feature $f_{\al}$ and all the edges between variables in $\bx_{\al}$ (if these variables are not also in the scope of other features) without changing the distribution of $\bx$. Therefore, by learning the parameters of this MRF model we can simultaneously learn the structure of a model if we allow some parameters to go to zero.

The Bayesian learning approach to graphical models considers parameters as a random variable subject to a prior. Given observed data, we can infer the posterior distribution of the parameters and their connectivity structure. Two commonly used priors, the Laplace and the Gaussian distribution, correspond to the $L_1$ and $L_2$ penalties respectively in the associated optimization-based approach. Although a model learned by $L_1$-penalized MLE is able to obtain a sparse structure, the full Bayesian treatment usually results in a fully connected model with many weak edges as observed in \citet{park2008bayesian}, without special approximate assumptions like the ones in \citet{lin2006bayesian}. We propose to use the ``spike and slab" prior to learn a sparse structure for MRFs in a fully Bayesian approach. The spike and slab prior \citep{mitchell1988bayesian,ishwaran2005spike} is a mixture distribution which consists of a point mass at zero (spike) and a widely spread distribution (slab):
\be
P(\ta_{\al}) = (1-p_0) \de(\ta_{\al}) + p_0 \mathcal{N}(\ta_{\al}; 0, \sg_0^2) \label{eqn:ss}
\ee
where $p_0\in [0,1]$, $\de$ is the Dirac delta function, and $\sg_0$ is usually large enough to be uninformative. The spike component controls the sparsity of the structure in the posterior distribution while the slab component usually applies a mild shrinkage effect on the parameters of the existing edges even in a highly sparse model. This type of \emph{selective} shrinkage is different from the global shrinkage imposed by $L_1/L_2$ regularization, and enjoys benefits in parameter estimation as demonstrated in the experiment section.

The Bayesian MRF with the spike and slab prior is formulated as follows:
\begin{align}
P(\bx|\bta) &= \frac{1}{Z(\bta)} \exp\left(\sum_{\al} \ta_\al f_\al(\bx_\al) \right) \nn \\
\ta_\al &= Y_\al A_\al \nn \\
Y_\al &\sim \mathrm{Bern}(p_0) \quad p_0 \sim \mathrm{Beta}(a,b) \nn \\
A_\al &\sim \mathcal{N}(0,\sg_0^2) \quad \sg_0^{-2} \sim \Gamma(c, d) \label{eqn:bbm}
\end{align}
where $\bx$ is a set of state variables and $a$, $b$, $c$, and $d$ are hyper-parameters. In the experiments we will use pairwise features in which case $\al=(i,j)$ plus bias terms given by $\sum_i \ta_i f_i(x_i)$ in the expression for the log-probability. We will use a normal prior $\ta_{i} \sim \mathcal{N}(0,\sg_b^2)$ for these bias terms. $\sg_b$ is chosen to be large enough to act as an uninformative prior. $Y_\al$ is a binary random variable representing the existence of the edges in the clique $\bx_\al$, and $A_\al$ is the actual value of the parameter $\ta_\al$ when the edges are instantiated. It is easy to observe that given $p_0$ and $\sg_0$, $\ta_\al$ has the same distribution as in equation \ref{eqn:ss}. We use a hierarchical model for $\bta$ so that the inference will be insensitive to the choice of the hyper-parameters. In fact, experiments show that with a simple setting of the hyper-parameters, proper values of the sparsity parameter $p_0$ and the variance $\sg_0$ are learned automatically by our model for all the data sets without the necessity of cross-validation.

Unlike the optimization-based methods which estimate a single structure, the Bayesian approach expresses uncertainty about the existence of edges through its posterior distribution, $P(\mathbf{Y}|\mathcal{D})$. We have applied a simple thresholding on $P(Y_\al | \mathcal{D})$ for edge detection in the experiments although more sophisticated methods can conceivably give better results.

\begin{figure}[t!]
\centering
\includegraphics[width=.8\linewidth] {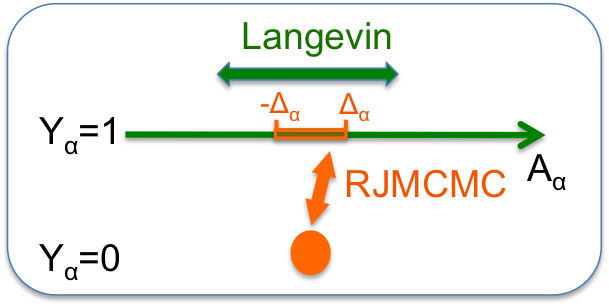}
\caption{\small Illustration of the MCMC for $A_{\al}$ and $Y_{\al}$.}
\label{fig:mcmc}
\end{figure}

Standard approaches to posterior inference do not work for Bayesian MRFs because it is intractable to compute the probability of a state $\bx$ (due to the intractability of the partition function). We devised an approximate MCMC algorithm for inference, where we draw samples of the continuous variable $A_{\al}$ by a modified Langevin dynamics algorithm, and samples of the discrete variable $Y_{\al}$ jointly with $A_{\al}$ by a reversible jump MCMC method, as illustrated in Figure \ref{fig:mcmc} and explained in the following sections.

\section{Sampling Parameter Values by Langevin Dynamics} \label{sec:lg}
Given $\mathbf{Y}$, $p_0$, $\sg_0$, and an observed data set $\mathcal{D}=\{\bx^{(m)}\}, m=1\dots N$, the
conditional distribution of parameters $\{A_{\al}: Y_{\al}=1\}$ is the posterior distribution of an MRF with a fixed edge set induced by $\{\al:Y_{\al}=1\}$ and an independent Gaussian prior $\mathcal{N}(0,\sg_0^2)$. We consider drawing samples of $A_{\al}$ with fixed $\mathbf{Y}$ in this section and will use $\ta_{\al}$ and $A_{\al}$ interchangeably to refer to a nonzero parameter. Even for an MRF model with a fixed structure, MCMC is still intractable. Approximate MCMC methods have been discussed in \citet{MurrayGhahramani04} among which Langevin Monte Carlo (LMC) with ``brief sampling" to compute the required expectations in the gradients, shows good performance. \citep{welling2011bayesian} further shows that LMC with a noisy gradient can draw samples from the exact posterior distribution when the step size approaches zero.

Langevin dynamics is described as the hybrid Monte Carlo (HMC) method with one leapfrog step in section 5.5.2 of \citet{neal2010mcmc}:
\begin{align}
\boldsymbol{p}_{t+\eps/2} =& \boldsymbol{p}_{t} + \frac{\eps C}{2} \boldsymbol{g} (\bta_{t}) \nn\\
\bta_{t+\eps} =& \bta_{t} + \eps C \boldsymbol{p}_{t+\eps/2} \nn\\
\boldsymbol{p}_{t+\eps} =& \boldsymbol{p}_{t+\eps/2} + \frac{\eps C}{2} \boldsymbol{g}(\bta_{t+\eps}) \label{eqn:lg_hmc}
\end{align}
where $\boldsymbol{p}$ is the auxiliary momentum, $\eps$ is the step size, $C$ is a positive definite preconditioning matrix, and $\boldsymbol{g}$ is the gradient of the log-posterior probability $\log P(\bta|\mathcal{D})$ \footnote{We omit all the other random variables that $\bta$ is conditioned on in this section for ease of notation.}. A new value of $\boldsymbol{p}$ is drawn at every iteration from an isotropic Gaussian distribution $\mathcal{N}(\mathbf{0},\mathbb{I})$ and then discarded after $\bta$ is updated. The leapfrog step is usually followed by a Metropolis-Hastings accept/reject step to ensure detailed balance. But since the rejection rate decays as $\eps^3$, that step can be skipped for small step sizes without incurring much error.

In a MRF, the gradient term $\boldsymbol{g}$ involves computing an expectation over exponentially many states as
\be
g_{\al}(\ta) = \sum_{m=1}^N f_{\al}(\bx_{\al}^{(m)}) - N \mathbb{E}_{P(\bx|\bta)}f_{\al}(\bx_{\al}) - \frac{\ta_{\al}}{\sg_{0}^2} \label{eqn:g}
\ee
The expectation is estimated by a set of state samples $\{\tilde{\bx}^{(s)}\}$ in the ``brief Langevin" algorithm of \citet{MurrayGhahramani04}, where these samples are drawn by running a few steps of Gibbs sampling initialized from a subset of the training data. We adopt the ``brief Langevin" algorithm with three modifications for faster mixing.

\subsection{Persistent Gibbs Sampling}
We maintain a set of persistent Markov chains for the state samples $\{\tilde{\bx}^{(s)}\}$ by initializing Gibbs sampling at the last states of the previous iteration instead of the data. This is motivated by the persistent contrastive divergence algorithm of \citet{Tieleman08}. When $\bta$ changes slowly enough, the Gibbs sampler will approximately sample from the stationary distribution, even when allowed a few steps at every iteration.

\subsection{Preconditioning}
When the posterior distribution of $\{\ta_{\al}\}$ has different scales along different variables, the original LMC with a common step size for all $\ta_{\al}$'s will traverse the parameter space slowly. We adopt a preconditioning matrix $C$ to speed up the mixing, where $C$ satisfies $CC^T=H$ with $H$ is the Hessian matrix of $\log P(\bta_{\textnormal{MAP}}|\mathcal{D})$, computed as:
\be
H(\bta_{\textnormal{MAP}}) = \mathrm{Cov}_{P(\bx|\bta_{\textnormal{MAP}})}\boldsymbol{f}(\bx) + \sg_0^{-2}
\ee
This is reminiscent of the observed Fisher information matrix in \citet{girolami2011riemann} except that we use the MAP estimate with the prior. We approximate $H(\bta_{\textnormal{MAP}})$ by averaging over $H(\bta_t)$ during a burn-in period and estimate $\mathrm{Cov}_{P(\bx|\bta_t)}\boldsymbol{f}$ with the set of state samples from the persistent Markov chains. The adoption of a preconditioning matrix also helps us pick a common step size parameter $\eps$ suitable for different training sets.

\subsection{Partial Momentum Refreshment}
The momentum term $\bp$ in the leapfrog step represents the update direction of the parameter. Langevin dynamics is known to explore the parameter space through inefficient random walk behavior because it draws an independent sample for $\bp$ at every iteration. 
We can achieve a better mixing rate with the partial momentum refreshment method proposed in \citet{horowitz1991generalized}. When $\bp$ is updated at every step by:
\be
\boldsymbol{p}_{t} \leftarrow \al \boldsymbol{p}_{t} + \bt \bn_{t}
\ee
where $n_{t} \sim \mathcal{N}(\mathbf{0},\mathbb{I})$, and $\al$, $\bt$ satisfy $\al^2 + \bt^2 = 1$, the momentum is partially preserved from the previous iteration and thereby suppresses the random-walk behavior in a similar fashion as HMC with multiple leapfrog steps. 

\begin{figure}[tb!]
\begin{minipage}[t]{.48\linewidth}
\centering
\includegraphics[width=\linewidth] {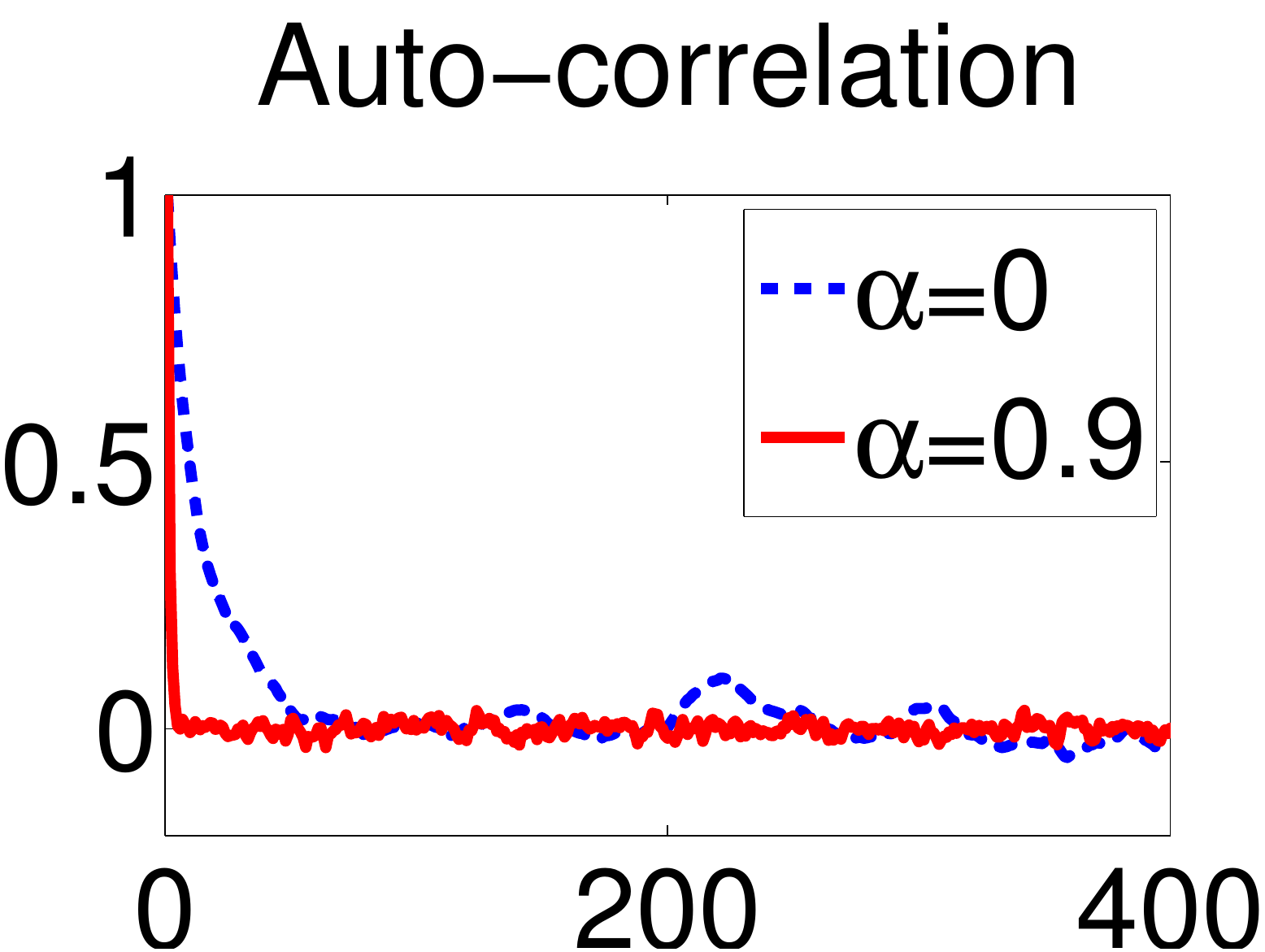}
\end{minipage}
\begin{minipage}[t]{.48\linewidth}
\centering
\includegraphics[width=\linewidth] {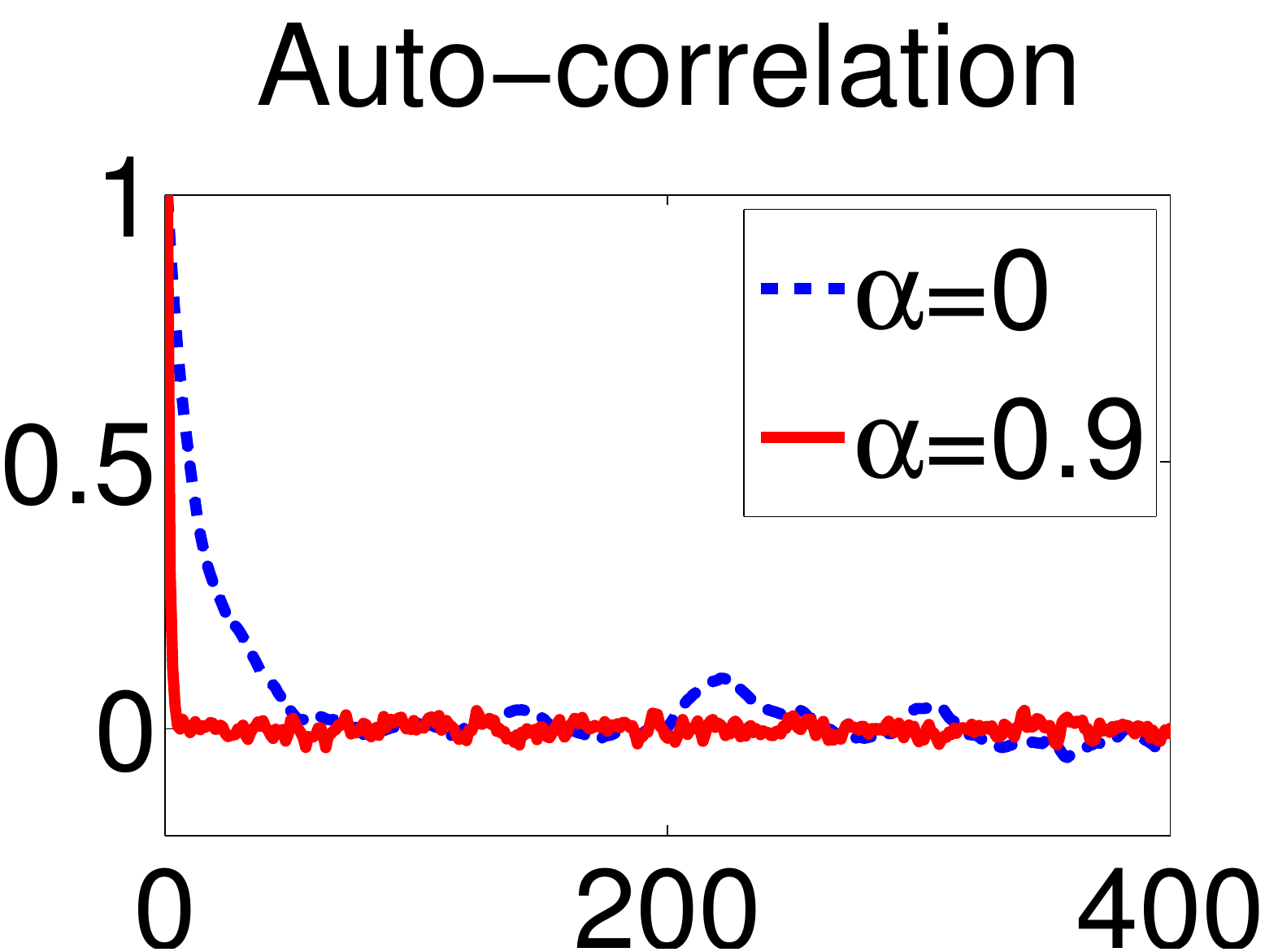}
\end{minipage}
\begin{minipage}[t]{.48\linewidth}
\centering
\includegraphics[width=\linewidth] {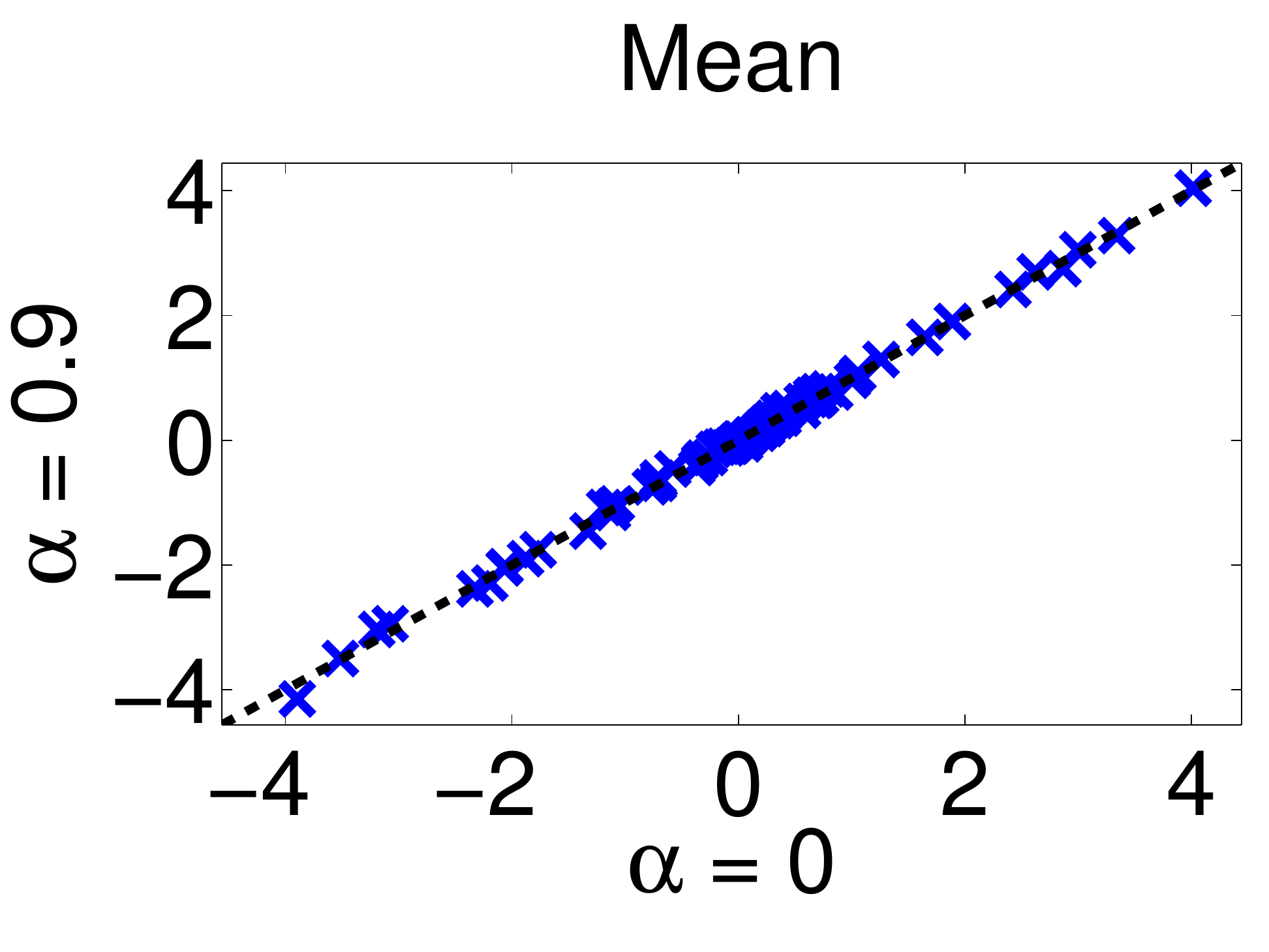}
\end{minipage}
\begin{minipage}[t]{.48\linewidth}
\centering
\includegraphics[width=\linewidth] {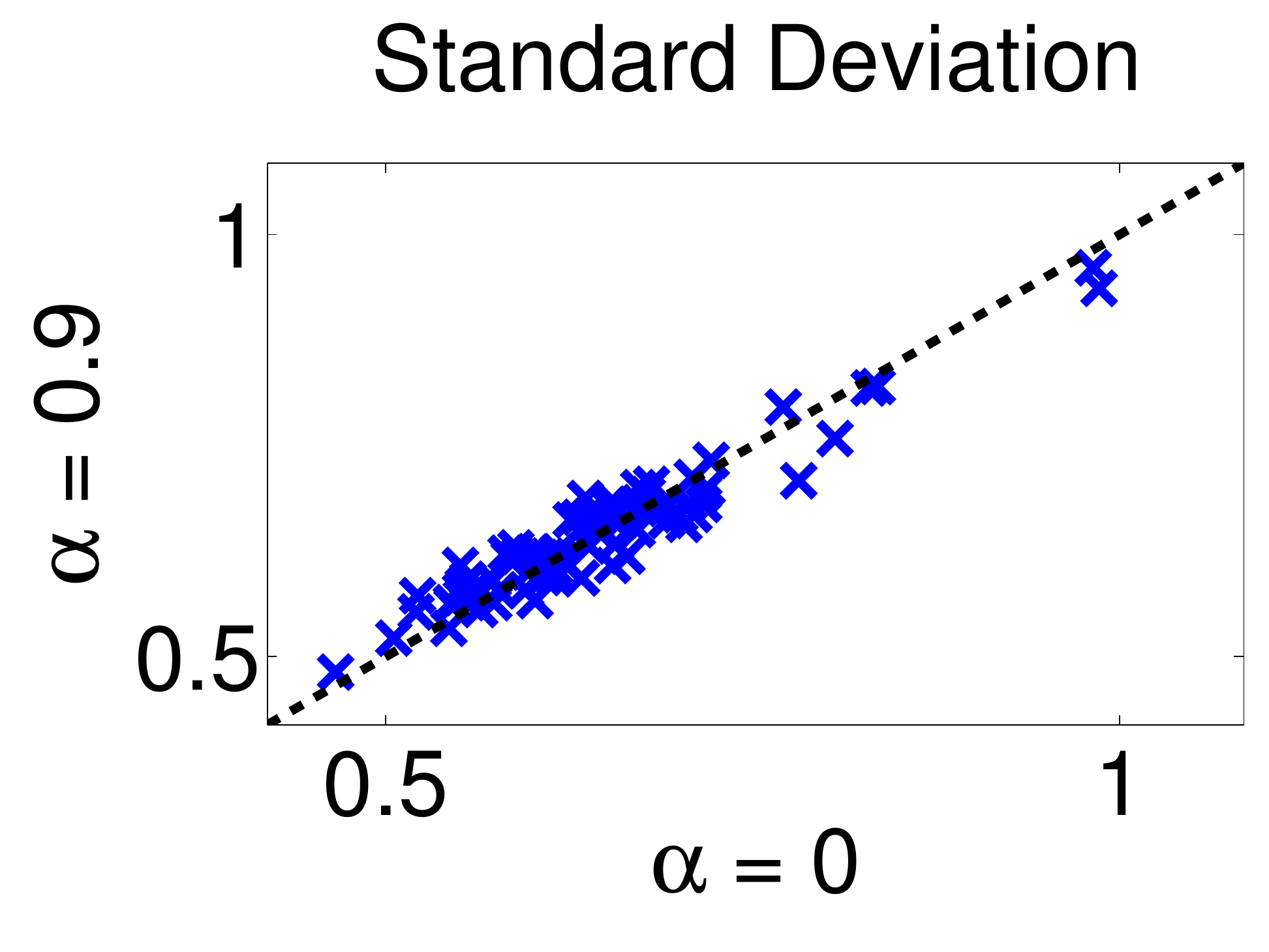}
\end{minipage}
\caption{\small Comparison of Langevin dynamics on the block model in section \ref{sec:dataset} with partial momentum refreshment $\al=0.9$ against $\al=0$. Step size $\eta=10^{-3}$. Top: the auto-correlation of two typical parameters. Bottom: the posterior mean and standard deviation of all parameters estimated with 10K samples.}
\label{fig:al_p}
\end{figure}

$\al$ controls how much momentum to be carried over from the previous iteration. With a large value of $\al$, LMC  reduces the auto-correlation between samples significantly relative to LMC without partial momentum refreshment. The improved mixing rate is illustrated in Figure \ref{fig:al_p}. We also show that the mean and standard deviation of the posterior distribution does not change. However, caution should be exercised especially when the step size $\eta$ is large because a value of $\al$ that is too large would increase the error in the update equation which we do not correct with a Metropolis-Hastings step because that is intractable.

\section{Sampling Edges by Reversible Jump MCMC} \label{sec:rjmcmc}
Langevin dynamics handles the continuous change in the parameter value $A_{\al}$. As for discrete changes in the model structure, $Y_{\al}$, we propose an approximate reversible jump MCMC (RJMCMC) step \citep{green1995reversible} to sample $Y_{\al}$ and $A_{\al}$ jointly from the conditional distribution $P(\mathbf{A}, \mathbf{Y}|p_0,\sg_0,\mathcal{D})$. The proposed Markov chain adds/deletes one clique (or simply one edge when $\al=(i,j)$) at a time. When an edge does not exist, i.e., $Y_{\al}=0$, the variable $A_{\al}$ can be excluded from the model, and therefore we consider the jump between a full model with $Y_{\al}\neq 0$, $A_{\al}=a$ and a degenerate model with $Y_{\al}=0$.

The proposed RJMCMC is as follows: when $Y_{\al}=0$, propose adding an edge with probability $P_{add}$ and sample $A_{\al}=a$ from a proposal distribution $q(A)$ with support on $[-\Delta_{\al},\Delta_{\al}]$; when $Y_{\al}=1$ and $|A_{\al}| \leq \Delta_{\al}$, propose deleting an edge with $P_{del}$. The reason of restricting the proposed move within $[-\Delta_{\al},\Delta_{\al}]$ will be explained later. It is easy to see that the Jacobian is $1$. The jump is then accepted by the Metropolis-Hastings algorithm with a probability:
\begin{align}
Q_{add} =& \min\{1, Q^*(a)\},\quad Q_{del} = \min\{1, 1/Q^*(a)\} \nn\\
Q^*(a) =& \exp(a \sum_m f_{\al}(\bx_{\al}^{(m)})) \left(\frac{Z(Y_{\al}=0)}{Z(Y_{\al}=1, A_{\al}=a)}\right)^N \nn\\
&  \frac{p_0 \mathcal{N}(A_{\al}=a|0,\sg_0^2)}{(1-p_0)} \frac{P_{del}}{P_{add} q(A_{\al}=a)} \label{eqn:rj_Q_add}
\end{align}
The factors in the first line of $Q^*$ represent the ratio of the model likelihoods while the other two are respectively the ratio of the prior distributions and the ratio of the proposal distributions.

\subsection{Unbiased Estimate to $Q^*$ and $1/Q^*$}
Computing the partition functions in equation \ref{eqn:rj_Q_add} is generally intractable. However, noticing that $Z(Y_{\al}=1, A_{\al}=a) \rightarrow Z(Y_{\al}=0)$, as $a\rightarrow 0$, the log-ratio of the two partition functions should be well approximated by a quadratic approximation at the origin when $a$ is small.  In this way we reduce the problem of estimating the partition function to computing the first and second order derivatives of the log-partition function. We employ a second order Taylor expansion for the ratio of partition functions in $Q^*$ as follows:
\begin{align}
&\left(\frac{Z(Y_{\al}=0)}{Z(Y_{\al}=1, A_{\al}=a)}\right)^N \stackrel{\text{def}}{=} R_{add} \nn\\
&\approx  \exp\left(- N a \frac{\partial \log(Z)}{\partial A_{\al}}\mid _{A_{\al}=0} - \frac{N a^2}{2} \frac{\partial^2 \log(Z)}{\partial A_{\al}^2}\mid _{A_{\al}=0} \right) \label{eqn:taylor}
\end{align}
We know that the $k$th order derivatives of the log-partition function of an MRF correspond to the $k$th order centralized moments (or cumulants) of the features, that is,
\be
\frac{\partial \log(Z)}{\partial A_{\al}} = \mathbb{E}f_{\al},\quad\frac{\partial^2 \log(Z)}{\partial A_{\al}^2} = \mathrm{Var}f_{\al} \label{eqn:derivative}
\ee
Given a set of $n$ state samples $\tilde{\bx}^{(s)}\sim P(\bx|Y_{\al}=0)$ from the persistent Markov chains, we can compute an unbiased estimate of the mean and variance of $f_{\al}$ as the sample mean $\bar{f}_{\al}$ and sample variance $S_{\al}^2=\sum_s(f_{\al}(\tilde{\bx}^{(s)})-\bar{f}_{\al})^2/(n-1)$ respectively. Consequently we obtain an estimate of $R_{add}$ by plugging $\bar{f}_{\al}$ and $S_{\al}^2$ into equation \ref{eqn:taylor}, which is unbiased in the logarithmic domain, denoted as $\hat{R}_{add}$. 

An unbiased estimate in the logarithmic domain, unfortunately, is no longer unbiased once transformed to the linear domain. To correct the bias induced by the transformation, we take another Taylor expansion of $\hat{R}_{add}/R_{add}$ around $a=0$. After some derivation, we obtain an unbiased estimate of $R_{add}$ upto the second order of $a$ given by
\be
\tilde{R}_{add}(a) = \exp\left(- N a \bar{f}_{\al} - \frac{N a^2}{2} S_{\al}^2 \right)  \left(1+\frac{N^2a^2}{2n}S_{\al}^2\right)^{-1} \nn\\
\ee
with variance:
\be
\mathrm{Var}(\tilde{R}_{add}/R_{add}) = \frac{N^2 a^2}{n}\mathrm{Var}(f_{\al}) + o(a^3) \label{eqn:rj_R_var}
\ee
Similarly, we can also obtain an unbiased estimate, $\tilde{R}_{del}$, in $1/Q^*$ when considering deleting an edge, with the same formula as $\tilde{R}_{add}$ except that $a$ is replaced by $-a$ and the sample mean and variance are now estimated with respect to $P(\bx|Y_{\al}=1, A_{\al}=a)$. If we plug in $\tilde{R}_{add}$ (or $\tilde{R}_{del}$) into $Q_{add}$ (or $Q_{del}$) we get an unbiased estimate of the acceptance probability except when $Q^*$ (or $1/Q^*$) is close to $1$ in which case the $\min$ operation causes extra bias. Since the variance can be computed as a function of $a$ in equation \ref{eqn:rj_R_var}, we can estimate how large a jump range $\Delta$ can be used in order to keep the error in the acceptance probability negligible. A larger data set requires a smaller jump range or alternatively a larger sample set that grows quadratically with the size of the data set.

\subsection{Optimal Proposal Distribution $q$}
After plugging equations \ref{eqn:taylor}, \ref{eqn:derivative} and \ref{eqn:g} into equation \ref{eqn:rj_Q_add}, we obtain the acceptance probability as a function of $a$:
\begin{align}
&Q_{add} = \min\left\{1, \frac{h(a)}{q(A_{\al}=a)}\mathrm{const}\right\} \nn\\
&h(a) = \exp\left(-\frac{a^2}{2}\left(\frac{1}{\sg_0^2} + \mathrm{Var}f_{\al}\right) + a Ng_{\al}\right)
\end{align}
where $g_{\al}$ and $\mathrm{Var}f_{\al}$ are defined at $\ta_{\al}=0$. Clearly, the optimal proposal distribution in terms of minimal variance is given by the following truncated Gaussian distribution
\be
q_{opt}(A_{\al}=a|\bta) \propto h(a), \quad a\in [-\Delta_{\al},\Delta_{\al}]
\ee
When adding an edge, we have state samples $\tilde{\bx}^{(s)}\sim P(\bx|Y_{\al}=0)$. The expectation $\mathbb{E}f_{\al}$ in $g_{\al}$ can be estimated by $\bar{f}_{\al}(\{\tilde{\bx}^{(s)}\})$ and $\mathrm{Var}f_{\al}$ by $S_{\al}^2(\{\tilde{\bx}^{(s)}\})$. When deleting an edge, the samples are from $P(\bx|Y_{\al}=1, A_{\al}=a)$. We apply a quadratic approximation for $\log Z(\ta_{\al})$, use equation \ref{eqn:derivative}, and estimate $\mathrm{Var}f_{\al} \approx S_{\al}^2$ and $\mathbb{E}f_{\al} \approx \bar{f}_{\al} - a S_{\al}^2$.

\subsection{Parallel Proposals}
Since the most computationally demanding step is to obtain a set of state samples $\{\tilde{\bx}^{(s)}\}$, we want to reuse the samples whenever possible. Given that $\Delta_{\al}$ is small enough, the parameter value does not change much when we accept an ``add or delete edge" move. We can thus assume that the distribution of $A_{\al}$ on an edge is not affected much by an accepted move on other edges. As a result we do not have to rerun the Gibbs sampler after every edge operation, and we can propose jumps for all $\{\al: |A_{\al}|<\Delta_{\al}\}$ in parallel, using the same set of samples. This reduces the computation time significantly.

\section{Sampling for Hyper-parameters} \label{sec:others}
Given $\mathbf{A}$ and $\mathbf{Y}$, the hyper-parameters are easy to sample when using conjugate priors:
\begin{align}
&p_0|\mathbf{Y} \sim \mathrm{Beta}(a + \sum_{\al}I(Y_{\al}=1), b + \sum_{\al}I(Y_{\al}=0)) \nn\\
&\sg_0^{-2} | \mathbf{Y},\mathbf{A} \sim \Gamma(c + \frac{1}{2} \sum_{\al}I(Y_{\al}=1), d + \frac{1}{2} \sum_{\al: Y_{\al}=1} A_{\al}^2)
\end{align}
where $I$ is the indicator function.

The whole inference algorithm is summarized in Alg \ref{alg:lvrjmcmc}.
\begin{algorithm}
\caption{MCMC for Bayesian MRFs with Langevin Dynamics and RJMCMC}
\label{alg:lvrjmcmc}
\begin{algorithmic}
\STATE (Parameters: number of iterations $ITER$, number of Gibbs sampling steps $N_{Gibbs}$, sample size $n$, step size $\eps$, partial momentum refreshment $\al$, RJMCMC proposal width $\Delta_{\al}$.)
\STATE Initialize $\mathbf{A}$, $\mathbf{Y}$, and momentum $\mathbf{p}$ randomly
\FOR{$iter = 1 \to ITER$} 
\STATE Sample $p_0$ and $\sg_0$ given $\mathbf{A}$ and $\mathbf{Y}$
\STATE Run Gibbs sampling for $N_{Gibbs}$ steps to draw $\{\bx^{(s)}\}_{s=1}^n$.
\STATE Run LMC to draw $\{A_{\al}:Y_{\al}\neq 0\}$.
\STATE Run RJMCMC to propose adding an edge for $\{\al: Y_{\al}=0\}$, and deleting an edge for $\{\al: Y_{\al}=1, |A_{\al}|<\Delta_{\al}\}$
\ENDFOR
\end{algorithmic}
\end{algorithm}

\section{Experiments} \label{sec:experiments}
\subsection{Datasets}\label{sec:dataset}
We assess the performance of our proposed method on two simulated data sets and one real-world data set. For the simulated data, we randomly generate sparse Ising models with binary $\pm 1$ states and with parameters sampled from a Gaussian distribution $\mathcal{N}(0,\sg^2)$ where $\sg=0.5$ for edges and $0.1$ for node biases. These models are then converted to their equivalent Boltzmann machines with binary $\{0,1\}$ states from which we draw exact samples. Two structures are considered: (1) a block model with $12$ nodes equally divided into $3$ groups. Edges are added randomly within a group with a probability of $0.8$, and across groups with $0.1$. Edges within a group are strong and positively coupled. There are $66$ candidate edges with $20$ edges in the ground truth model. (2) a $10\times 10$ lattice with $4950$ candidate edges and with $180$ edges in the ground truth model.

For the real data, we use the MNIST digits image data. We convert the gray scale pixel values to binary values by thresholding at a value of $50$, resize the images to a $14\times 14$ scale, and then pick a $9\times 12$ patch centered in each image where the average value of each pixel is in the range of $[10^{-4},1-10^{-4}]$. The last step is necessary because the other competing models do not have regularization on their biases, which will result in divergent parameter estimates for pixels that are always $0$ or $1$.

\subsection{Model Specification}
We compare our Bayesian structure learning algorithm with two other approaches. One is proposed in \citet{wainwright2007high} which recovers the neighbourhood of nodes by training separate $L_1$-regularized logistic regressions on each variable. While its goal is edge detection, we can also use it as a parameter estimation method with two output models, ``Wain Max" and ``Wain Min", as defined in \citet{höfling2009estimation}. We implement the ``Wain Max/Min" methods with the Lasso regularized generalized linear model package of \citet{friedman2010regularization} \footnote{Code provided at http://www-stat.stanford.edu/~tibs/glmnet-matlab}. The other approach is one of the several variants of $L_1$-regularized MLE methods which use a pseudo-likelihood approximation \citep{höfling2009estimation}, denoted as ``MLE" \footnote{Code provided at http://holgerhoefling.com}. 

For our Bayesian model, we consider two schemes to specify a model for prediction. One is the fully Bayesian approach, referred as ``Bayes", in which we random pick $100$ model samples in the Markov chain and approximate the Bayesian model by a mixture model of these $100$ components. The other one is to obtain a single model by applying a threshold of $0.5$ to $P(Y_{i,j}|\mathcal{D})$ and estimate the posterior mean of the included edges, referred to as ``Bayes PM".

The performance of the Bayesian model is insensitive to the choice of hyper-parameters. We simply set $a=b=c=d=5$ for $p_0$ and $\sg_0$, and $\sg_b=10$ across all experiments. For the parameters of the MCMC method, we also use a common setting. We use a diagonal approximation to the feature covariance $\mathrm{Cov}\boldsymbol{f}$ and thereby the preconditioning matrix $C$. We set the sample size $n=100$, number of Gibbs sampling steps $N_{Gibbs}=1$, LMC step size $\eps=10^{-3}$, and momentum refreshment rate $\al=0.9$. The RJMCMC proposal width is set as $\Delta_{\al}=0.01/\sqrt{N\mathrm{Var}f}$ to achieve a small estimation error as in equation \ref{eqn:rj_R_var}. For each experiment, we run the MCMC algorithm to collect $10K$ samples with a subsampling interval of $1000$. Since the exact partition function can be computed on the small block model, we also run an exact MCMC, ``Bayes Exact", with an exact gradient and accept/reject decision as well as larger values in $\eps$, $\al$, and $\Delta$ than the approximate MCMC.

Different levels of sparsity have to be considered in $L_1$-based methods for an optimal regularization strength. For the Bayesian method, we learn a single sparsity level. However, for the sake of comparison, we also consider a method with $p_0$ as a parameter and vary it between $(0,1)$ to induce different sparsity, referred to with a suffix ``$p_0$".

\subsection{Accuracy of Inference}
\begin{figure}[tb!]
\begin{minipage}[t]{\linewidth}
\centering
\includegraphics[width=.9\linewidth] {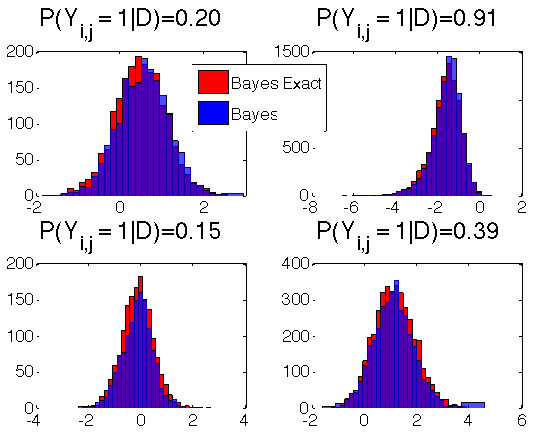}
\vspace{-.4cm}
\caption{\small Histogram of the parameter samples of four randomly picked edges from ``Bayes" and ``Bayes Exact" when $Y_{i,j}=1$. The training data is from the block model with $N=100$.}
\label{fig:validity_hist}
\end{minipage}
\begin{minipage}[t]{\linewidth}
\centering
\includegraphics[width=\linewidth] {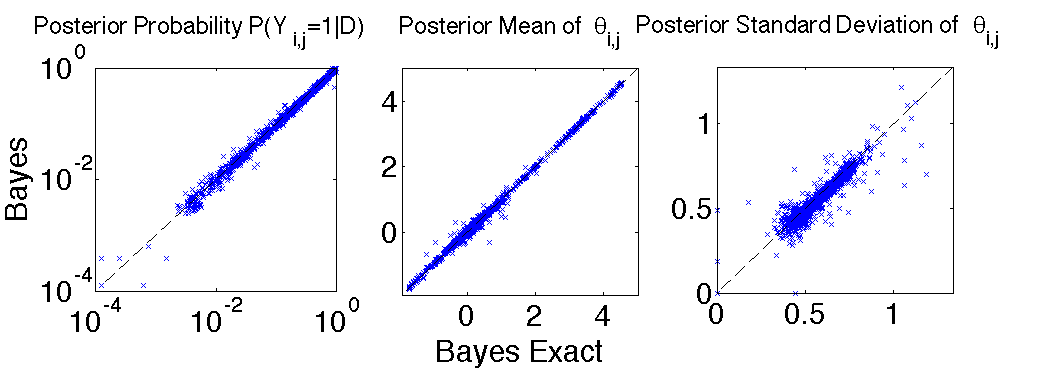}
\vspace{-.6cm}
\caption{\small Comparison of ``Bayes $p_0$" and ``Bayes Exact $p_0$" on posterior probability of $Y_{i,j}=1$, and the posterior mean and standard deviation of $\ta_{i,j}$ when $Y_{i,j}=1$. The training data is from the block model with $N=100$.}
\label{fig:validity_matching}
\vspace{-.3cm}
\end{minipage}
\end{figure}

We first evaluate the validity of the proposed MCMC method on the block data where exact inference can be carried out. The marginal posterior distribution of an edge parameter, $\ta_{i,j}$, is a mixture of a point mass at zero and a continuous component with a single mode. Figure \ref{fig:validity_hist} shows the histogram of samples in the continuous component of four randomly picked $\ta_{i,j}$'s. The title above each plot is the posterior probability of the edge $(i,j)$ being present in the model, i.e. $\ta_{i,j}\neq 0$ or $Y_{i,j}=1$. In this figure, the marginal distribution estimated from the approximate MCMC method matches the distribution from ``Bayes Exact" very well. For a more comprehensive comparison, we run ``Bayes $p_0$" and ``Bayes Exact $p_0$" methods, and vary the value of $p_0$ from $10^{-4}$ to $0.99$ to achieve different levels of sparsity. We estimate and collect across different $p_0$ values the posterior probability of $\ta_{i,j}\neq 0$, posterior mean and standard deviation of $\ta_{i,j}$ in the continuous component, as shown in Figure \ref{fig:validity_matching}. Each point represents a parameter under some value of $p_0$. We find the approximate MCMC procedure produces about the same values for these three statistics as the exact MCMC method.

\subsection{Simulated Data}
We then compare the performance of various methods on simulated data sets for two tasks: structure recovery and model estimation. The accuracy of recovering the true structure is measured by the precision and recall of the true edges. The quality of the estimated models is evaluated by the predictive performance on a held-out validation set. Since computing the log-likelihood of a MRF is in general intractable, we use the conditional log-likelihood (CLL) on a group of variables instead, which is a generalization of the conditional marginal log-likelihood in \citet{lee2006efficient}. For each data case in the validation set, we randomly choose a group of variables, which is all the variables in the block model and a $3\times 3$ grid in the lattice and MNIST models, and compute $\log P(\{x_i\}_{i\in group}|\{x_j\}_{j\not\in group})$. 

\begin{figure}[tb!]
\centering
\includegraphics[width=.75\linewidth] {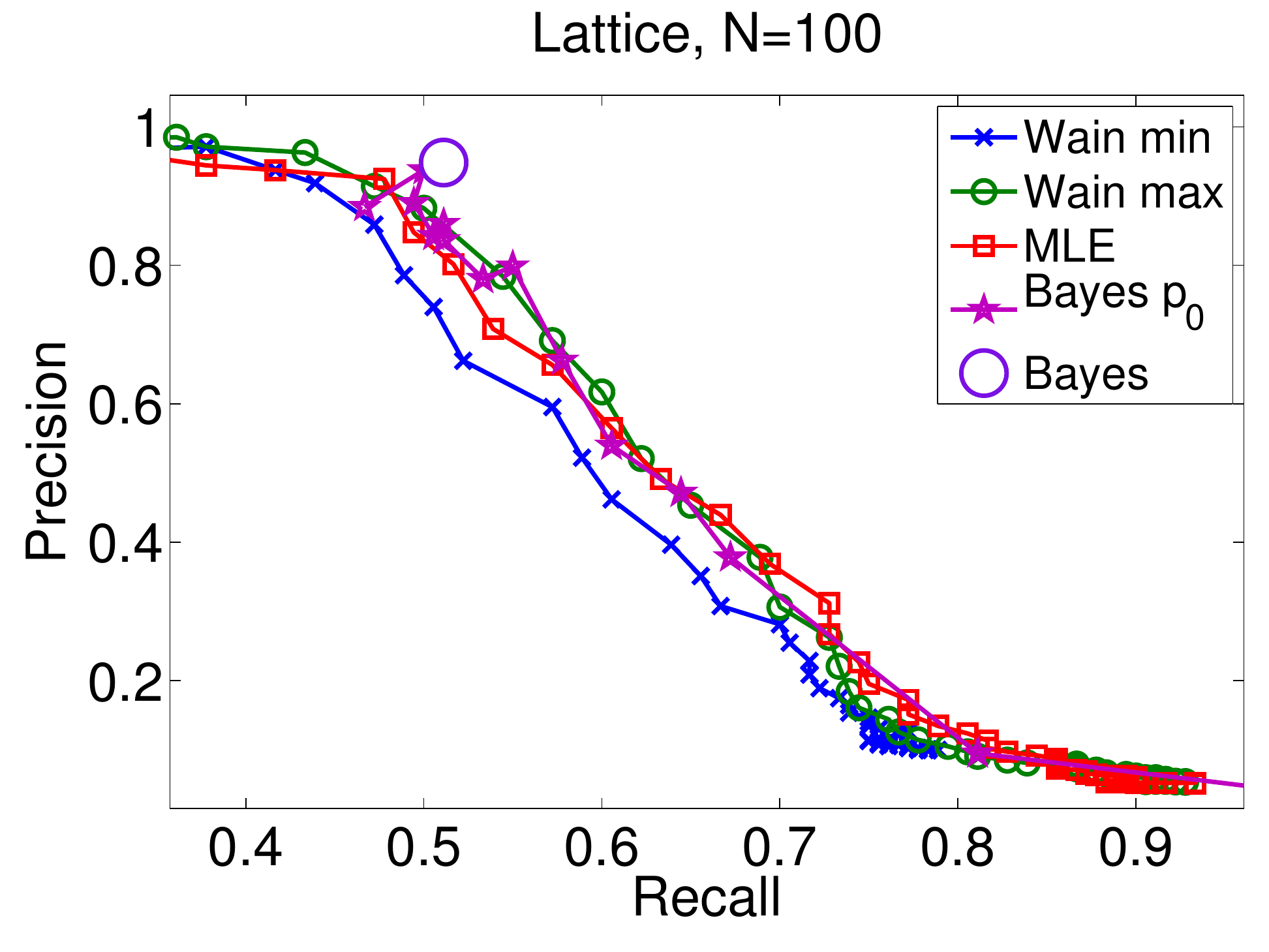}
\vspace{-.4cm}
\caption{\small Precision-Recall Curves for the lattice data with $100$ data cases.}
\label{fig:pr}
\vspace{-.3cm}
\end{figure}

We train models on the simulated data ranging between $50$ and $1000$ items. For the Bayesian models, we remove an edge if the posterior probability $P(Y_{i,j}=1|\mathcal{D})<0.5$. Figure \ref{fig:pr} shows typical precision-recall curves for different methods. It turns out that all the models with a sparsity tuning parameter perform similarly across all the training sets. The ``Bayes" model tends to find a structure with high precision.

\begin{figure*}[tb!]
\begin{minipage}[t]{.68\linewidth}
	\begin{minipage}[t]{.49\linewidth}
	\centering
	\includegraphics[width=\linewidth] {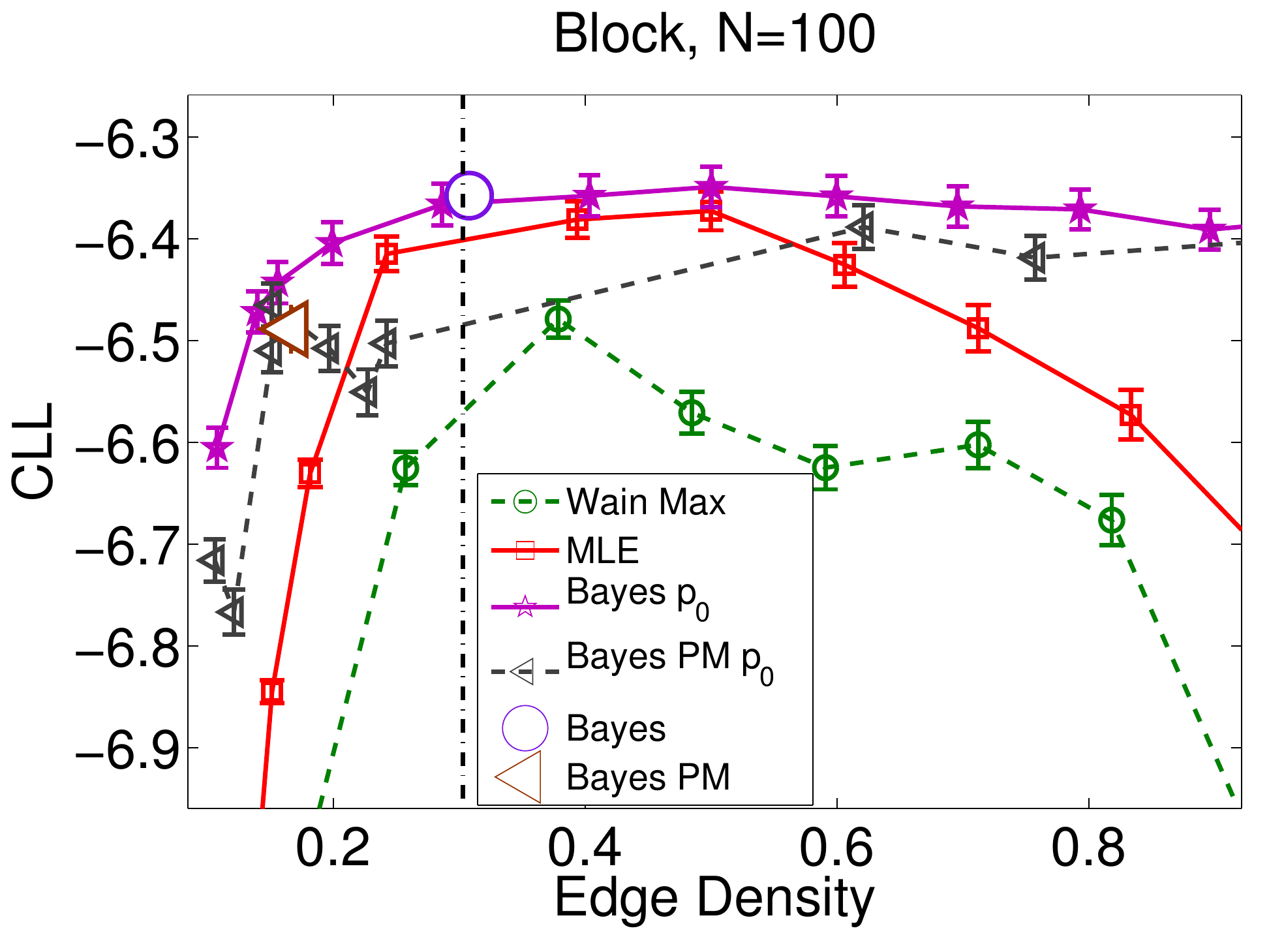}
	\end{minipage}
	\begin{minipage}[t]{.49\linewidth}
	\centering
	\includegraphics[width=\linewidth] {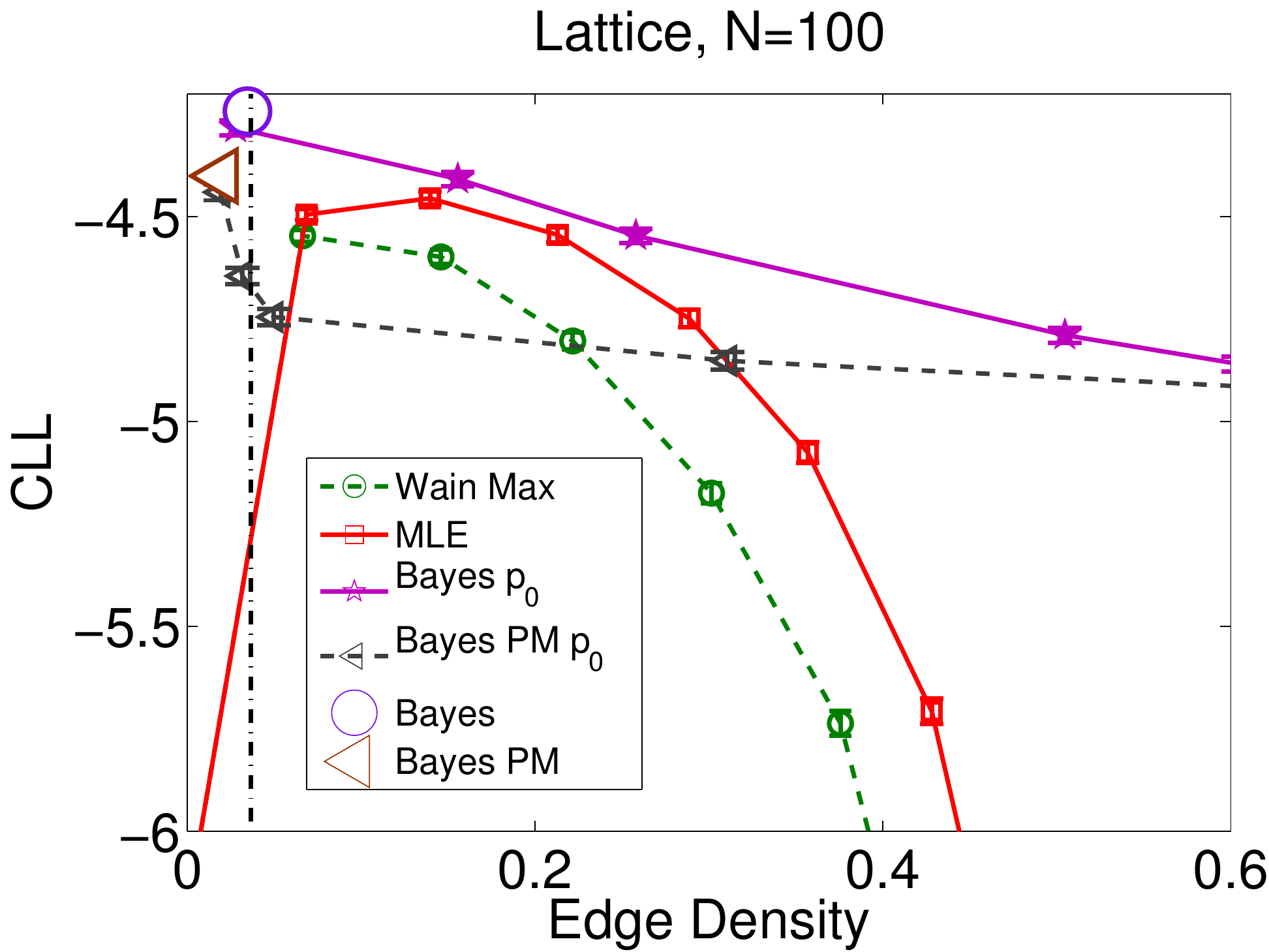}
	\end{minipage}
\vspace{-.3cm}
\caption{\small Mean and standard deviation of average CLL at different density levels for block (left) and lattice (right) model with $100$ training data cases. ``Wain Min" is not plotted as it is always inferior to ``Wain Max". Vertical line: true edge density.}
\label{fig:cll}
\end{minipage}
\hspace{.5cm}
\begin{minipage}[t]{.28\linewidth}
\centering
\includegraphics[width=\linewidth] {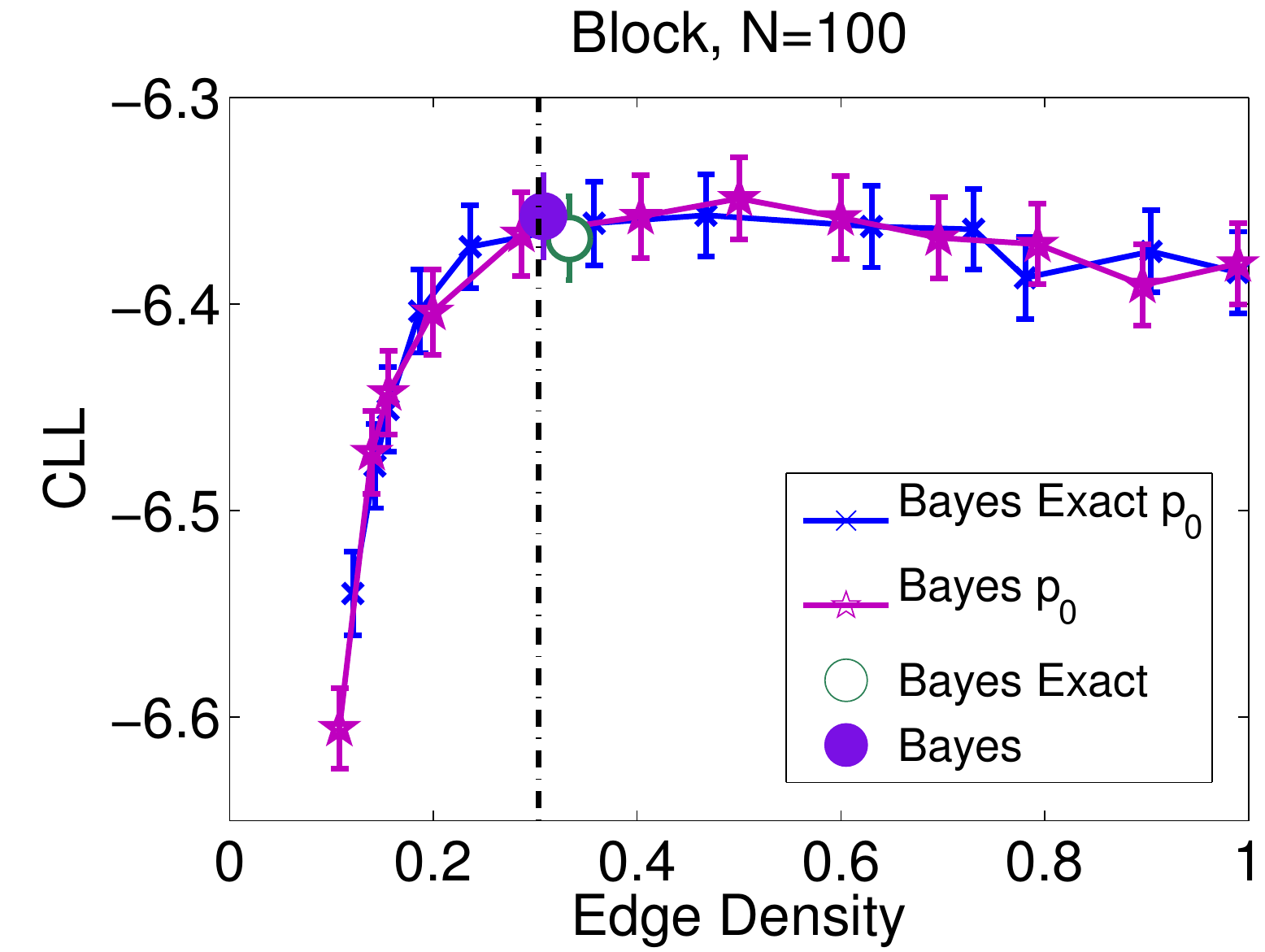}
\vspace{-.7cm}
\caption{\small Average CLL at different density levels for the block model with $100$ training cases.}
\label{fig:cll_validity}
\end{minipage}
\vspace{-.3cm}
\end{figure*}

We then consider the average CLL as a function of the edge density of a model. The edge density is defined as the percentage of edges present in a model, i.e., $1-sparsity$. For the fully Bayesian model, we measure the average density in the Markov chain. In fact, the variance of the edge density is usually small, suggesting that most model samples have about the same number of edges. We show the average CLL of different models trained with $100$ data cases in Figure \ref{fig:cll}. The results on other data sizes are omitted because they have the same tendency. All the Bayesian models give robust prediction performance in the sparse and dense model ranges and ``Bayes $p_0$" outperforms all the other methods with a tunable sparsity parameter for almost all settings of $p_0$. Moreover, the curve of ``Bayes $p_0$" peaks at the true density level. In contrast, $L_1$ methods would underfit to the data for sparse models or overfit for dense models. ``MLE" performs better than ``Wain" models. This makes sense as the ``Wain" models were not designed for MRF parameter estimation. Figure \ref{fig:cll_validity} compares the Bayesian models with exact and approximate inference. Again, the approximate MCMC method generates about the same posterior distribution as the ``exact" method.

\begin{figure}[tb!]
\vspace{-.3cm}
\centering
\includegraphics[width=\linewidth] {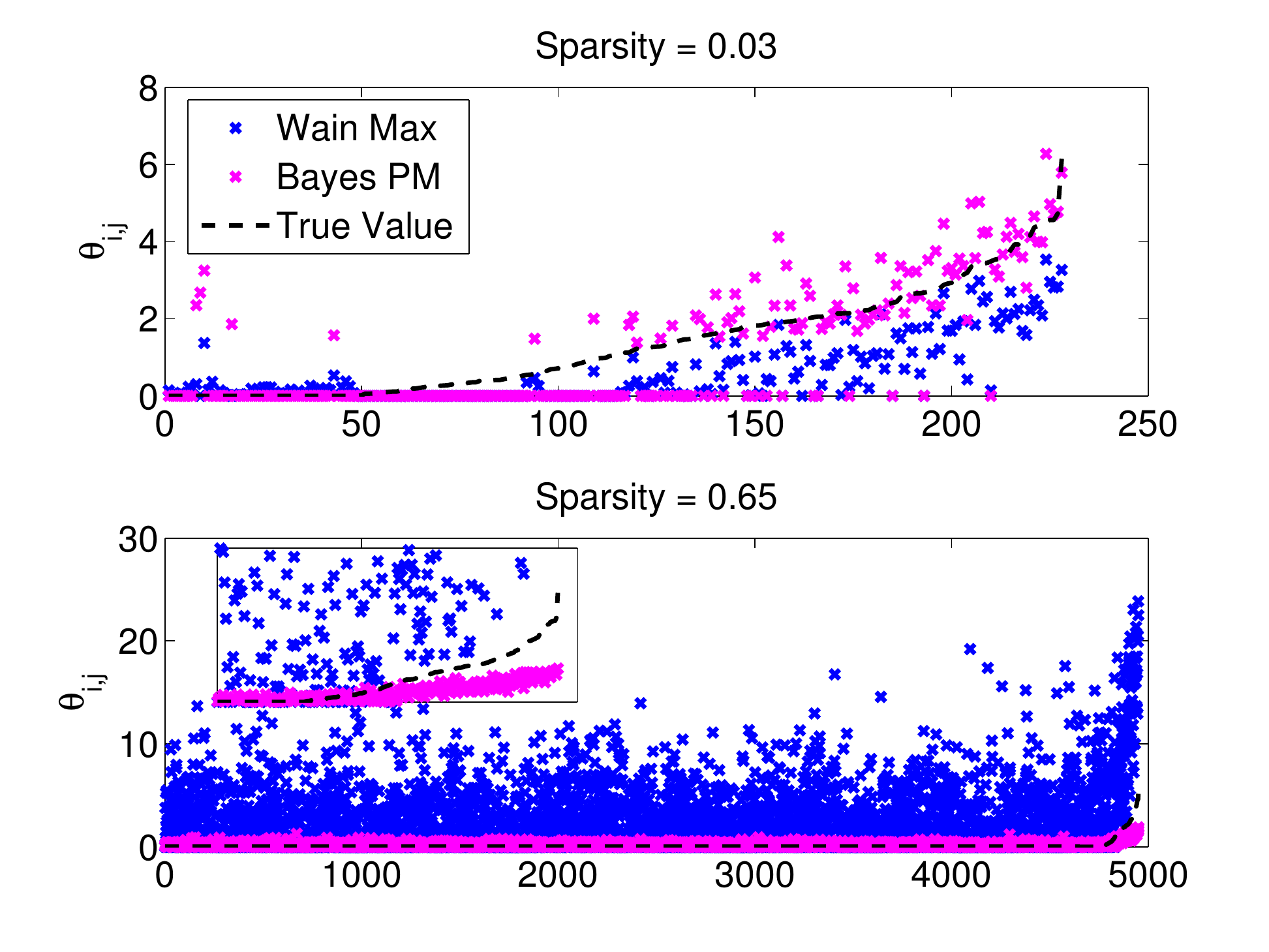}
\vspace{-.8cm}
\caption{\small Absolute value of edge parameters in the true model, ``Wain Max", and ``Bayes PM" for lattice data with $100$ samples. Parameters are plotted only if they are not zero in at least one model and sorted along the x-axis by the absolute value in the true model. The parameters in ``Wain Max'' are too small in sparse models (top, edge density $\approx$ $3\%$) and too large in dense models (bottom, $65\%$). Inset: zoom-in at the right-hand side. ``Wain Min" and ``MLE" are similar to ``Wain Max".}
\label{fig:underoverfit}
\vspace{-.5cm}
\end{figure}

The difference between Bayesian models and $L_1$-based models could be partially explained by the different prior/regularization. As shown in Figure \ref{fig:underoverfit}, to achieve a sparse structure, we have to use strong regularization in the $L_1$ models which causes global shrinkage for all parameters, resulting in under-fitting. On the other hand, to obtain a dense structure, weak regularization must be applied globally which leads to over-fitting. In contrast, with a spike and slab prior, the parameter value of existing edges is not affected directly by $p_0$. Instead, their variance is controlled by another random variable $\sg_0$ which fits the data automatically with a weak hierarchical prior. The behavior of selective shrinkage in the spike and slab prior is also discussed in \citet{mohamed2011bayesian,ishwaran2005spike}.

The Bayesian models, ``Bayes $p_0$" and ``Bayes PM $p_0$" do however show an interesting ``under-fitting" phenomenon at a large density levels. This is due to a misspecified value for $p_0$. Since the standard deviation $\sg_0$ is shared by all the $A_{i,j}$'s whose $Y_{i,j}=1$, when we fix $p_0$ at an improperly large value, it forces a lot of non-existing edges to be included in the model, which consequently brings down the posterior distribution of $\sg_0$. This results in too small values on real edges as shown in the inset of Figure \ref{fig:underoverfit} and thereby a decrease in the model predictive accuracy. 

\begin{figure}[tb!]
 \begin{minipage}[t]{\linewidth}
 \centering
 \includegraphics[width=.9\linewidth] {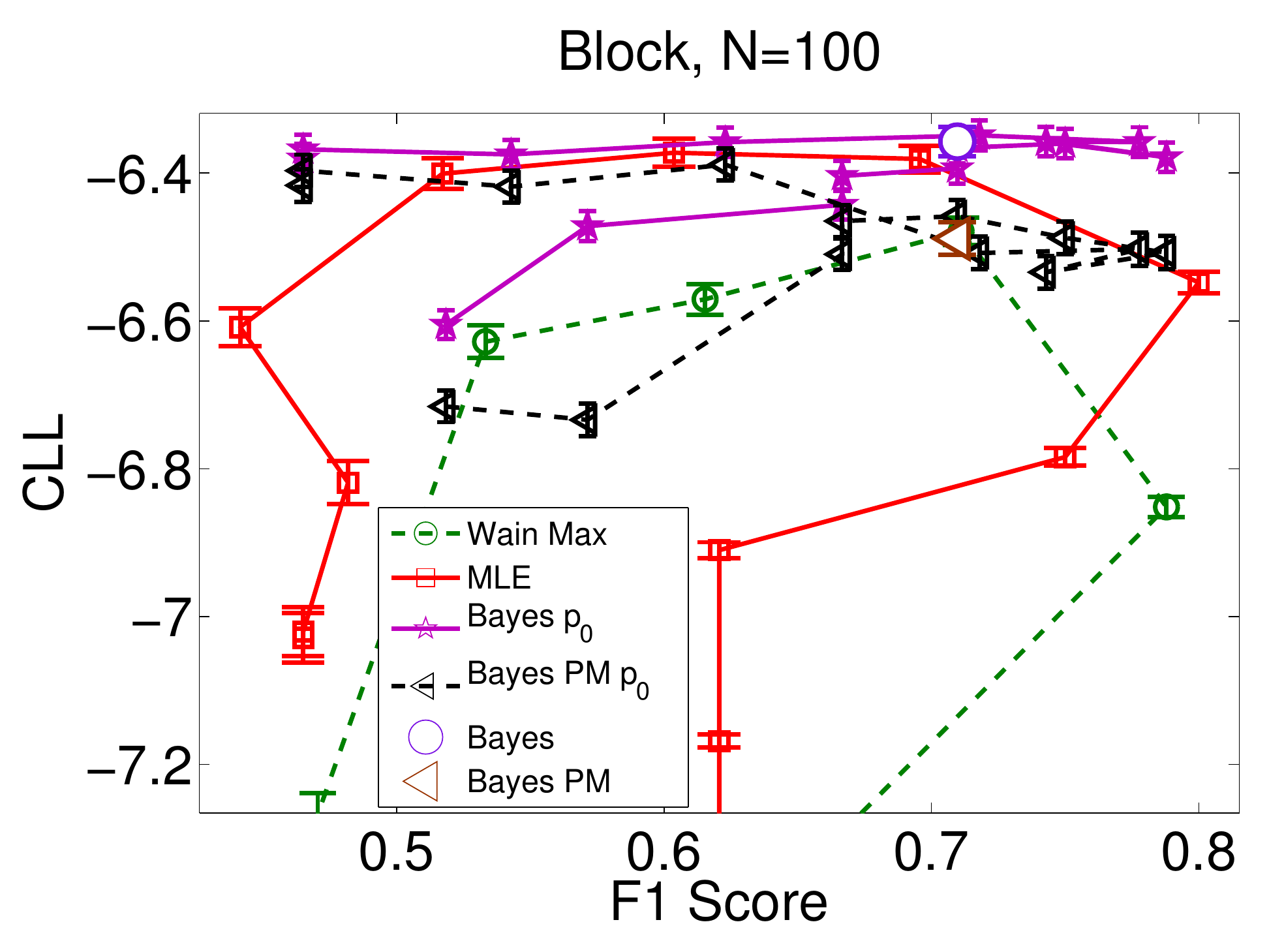}
 \end{minipage}
 \begin{minipage}[t]{\linewidth}
 \centering
 \includegraphics[width=.9\linewidth] {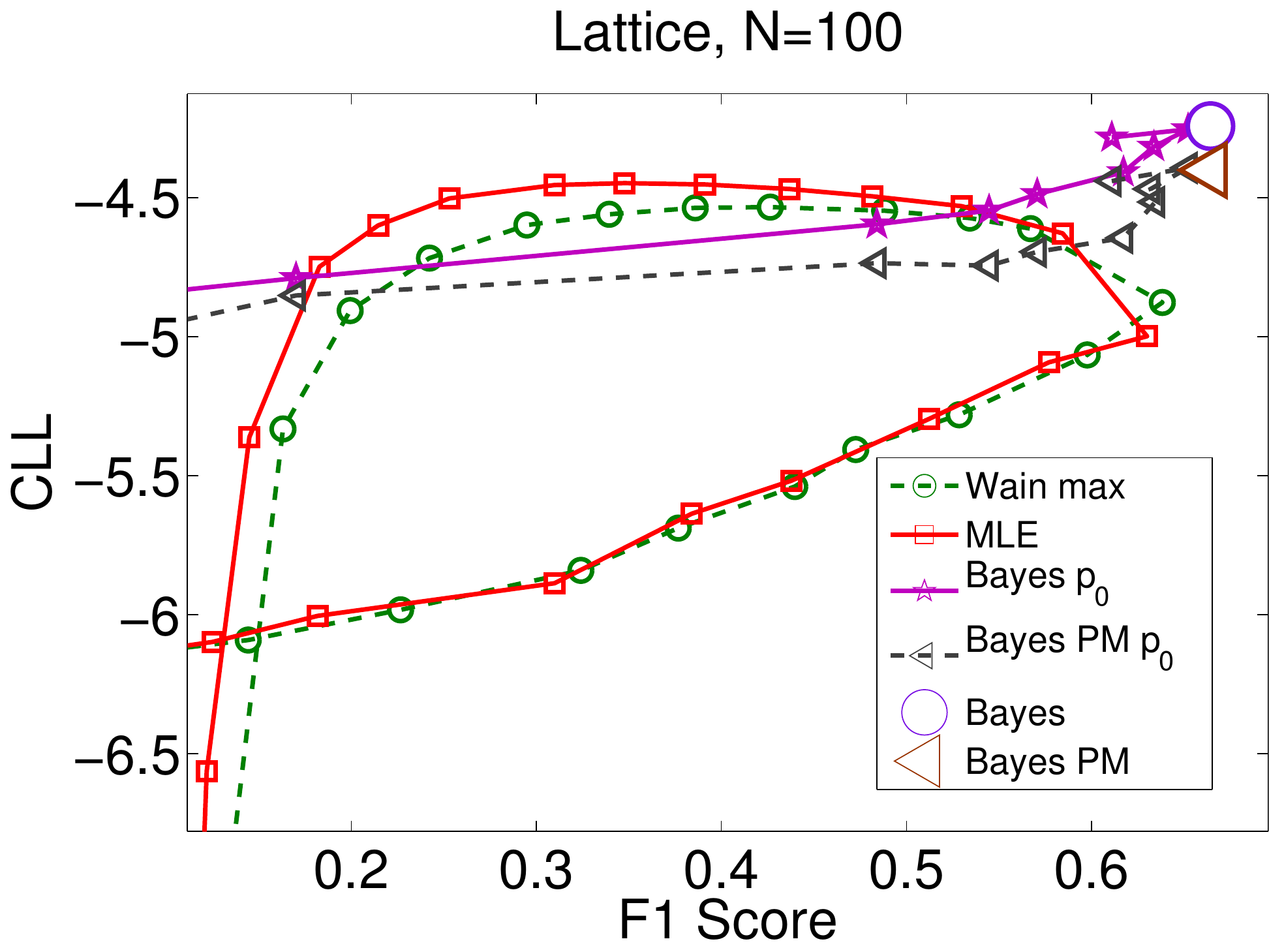}
 \end{minipage}
\vspace{-.8cm}
\caption{\small CLL vs F1 score for block and lattice model with $100$ data cases. A good model is at the upper right corner.}
\label{fig:f1cll}
\vspace{-.2cm}
\end{figure}

However, this ``misbehavior" in return just suggests the ability of our Bayesian model to learn the true structure. Once we release $p_0$ through a hierarchical prior, the model will abandon these improper values in $p_0$ and automatically find a good structure and parameters. The vertical line in Figure \ref{fig:cll} indicates the sparsity of the true model. Both ``Bayes" and ``Bayes PM" find sparsity levels very close to the true value, while $L_1$-based methods are under-fitted at that same level as shown in the upper panel of Figure \ref{fig:underoverfit}. We show the joint performance of edge detection and parameter learning in Figure \ref{fig:f1cll} where the performance of edge detection is summarized by the F1 score (the harmonic mean of the precision and recall). The Bayesian models with a hierarchical $p_0$ prior achieve both a high F-1 score and CLL value near the upper right corner.

\subsection{MNIST Data}
\begin{figure}[tb!]
	\begin{minipage}[t]{\linewidth}
	\centering
	\includegraphics[width=.9\linewidth] {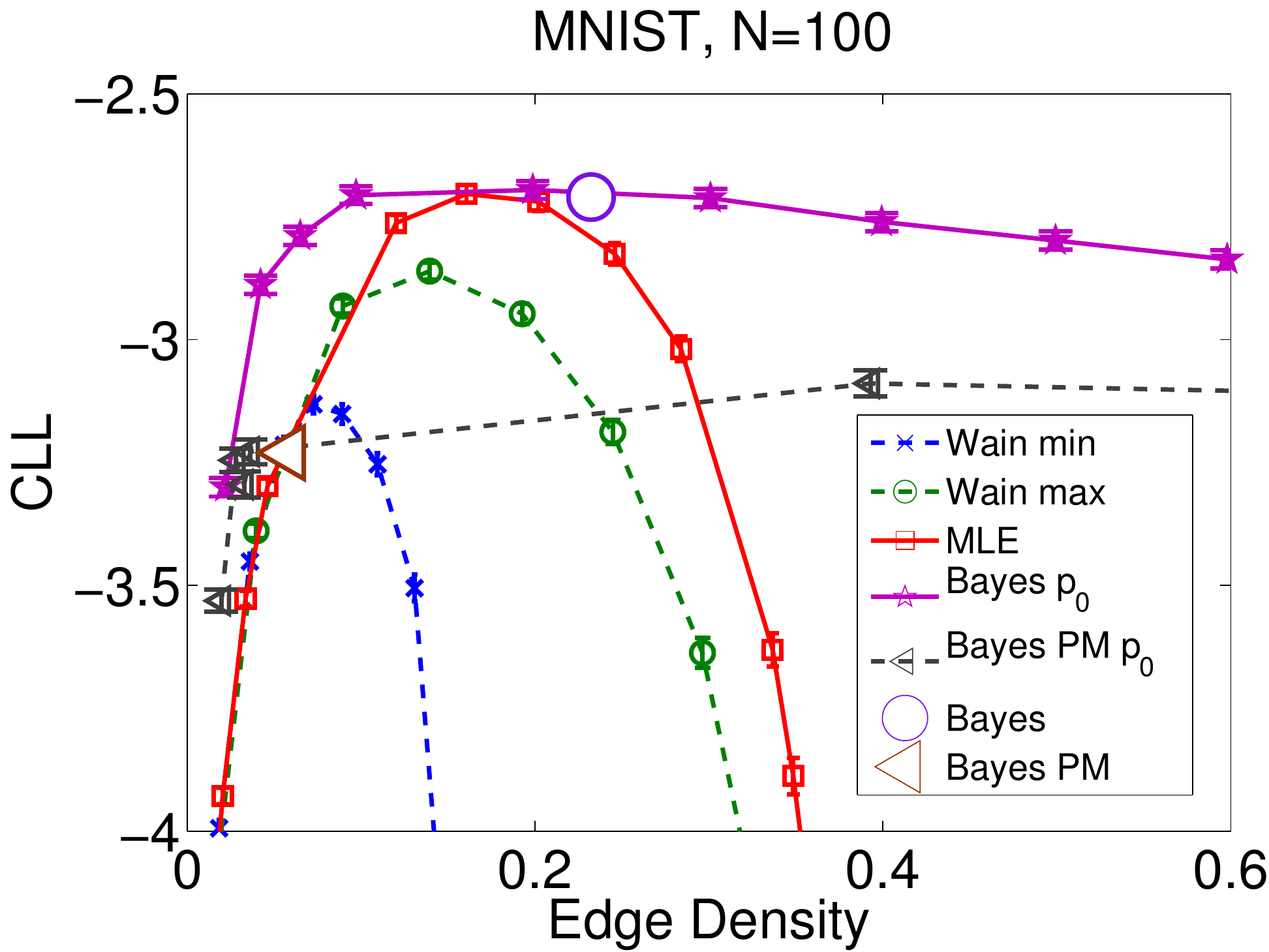}
	\end{minipage}
	\begin{minipage}[t]{\linewidth}
	\centering
	\includegraphics[width=.9\linewidth] {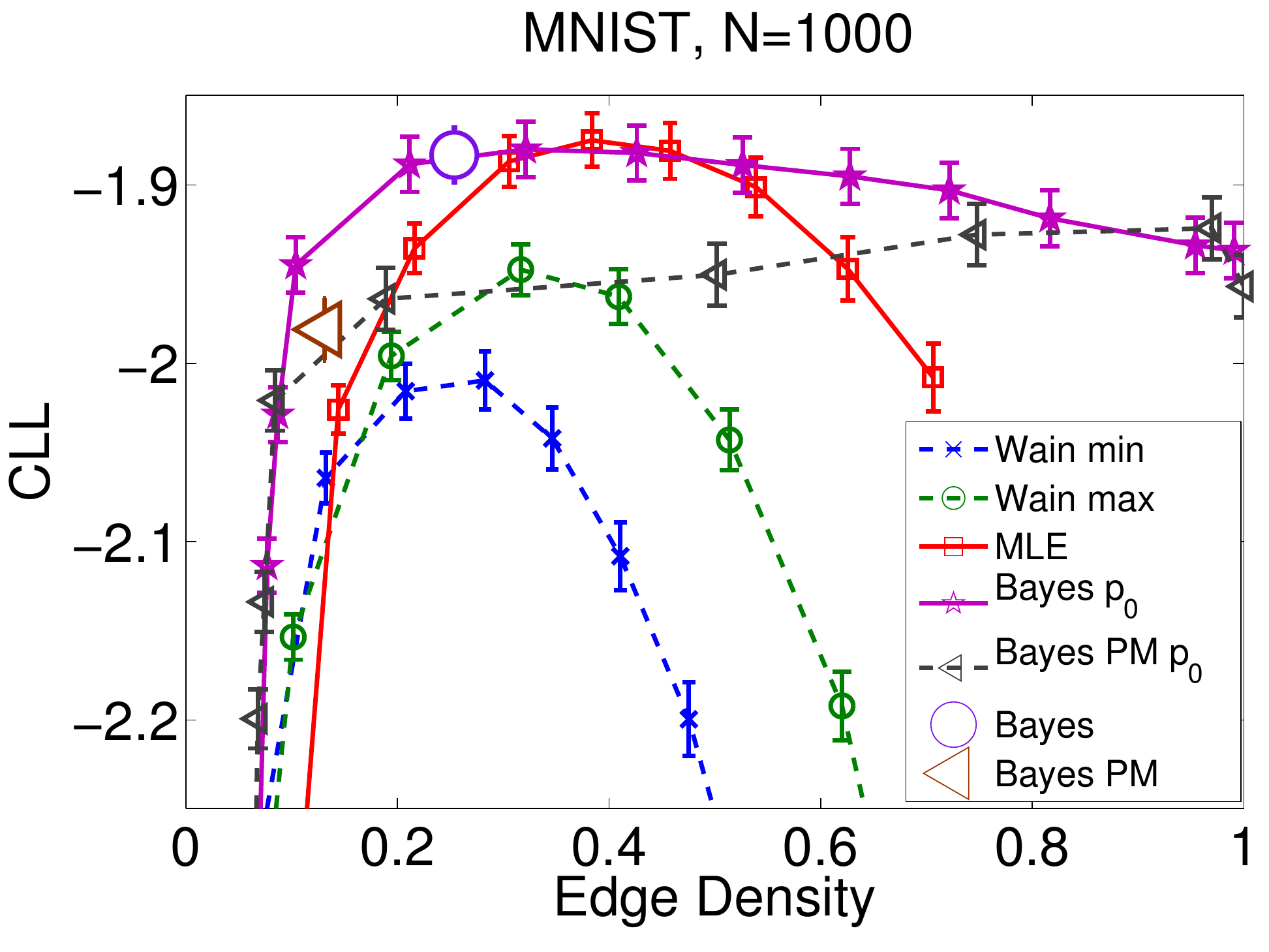}
	\end{minipage}
\vspace{-.8cm}
\caption{\small Mean and standard deviation of average CLL versus edge density on MNIST with $100$ and $1000$ data cases.}
\label{fig:mnist_cll}
\vspace{-.5cm}
\end{figure}

Since there does not exist ground truth in the model structure of the MIST data set, we evaluate how well we can learn a sparse model for prediction. Figure \ref{fig:mnist_cll} shows the average CLL on $10K$ test images with a model trained on $100$ and $1000$ images respectively. In the sparse and dense model ranges, we observe again a better performance of ``Bayes" than $L_1$-based methods. ``Bayes PM" also shows robustness to under/over-fitting although it seems that simply computing the posterior mean does not provide sufficiently good model parameters in the median density range.

\begin{figure}[tb!]
\centering
\includegraphics[width=\linewidth] {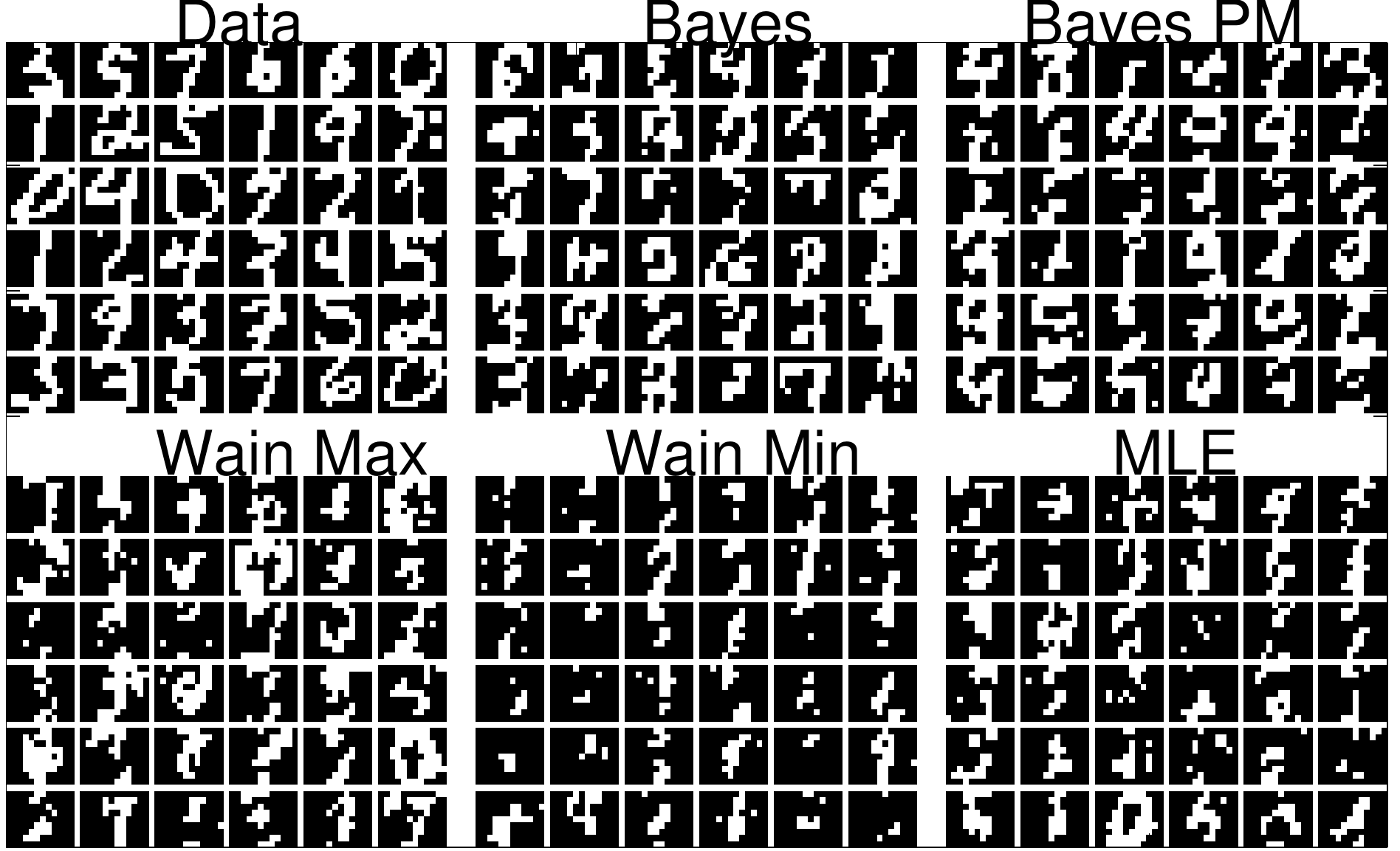}
\vspace{-.7cm}
\caption{\small Samples from learned models at an edge density level of $0.2$.}
\label{fig:mnist_sample}
\vspace{-.3cm}
\end{figure}

To get a more intuitive comparison about the quality of learned sparse models, we train models on $1000$ images  by different methods with a density of $0.2$ and then run Gibbs sampling to draw $36$ samples from each model. The images are shown in Figure \ref{fig:mnist_sample}. While it is hard to get good reconstruction using a model without hidden variables, the Bayesian methods produce qualitatively better images than competing methods, even though ``Bayes PM" does not have higher CLL than ``MLE" at this level.

A common limitation of learning Bayesian models with the MCMC inference method is that it is much slower than training a model with a point estimate. However, as shown in the experiments, the Bayesian methods are able to learn a good combination of parameters and a structure without the need to tune hyper-parameters through cross validation. Also, the Bayesian methods learn sparser models than $L_1$-based methods without sacrificing predictive performance significantly. Because the computational complexity of inference grows exponentially with the maximum clique size of MRFs, $L_1$-based models at their optimal (not so sparse) regularization level can in fact become significantly more computationally expensive than their Bayesian counterparts at prediction time. Turning up the regularization will result in sparser models but at the cost of under-fitting the data and thus sacrificing predictive accuracy.

\section{Discussion} \label{sec:discussion}
We propose Bayesian structure learning for MRFs with a spike and slab prior. An approximate MCMC method is proposed to achieve effective inference based on Langevin dynamics and reversible jump MCMC. As far as we known this is the first attempt to learn MRF structures in the fully Bayesian approach using spike and slab priors. Related work was presented in \citet{parise2006structure} with a variational method for Bayesian MRF model selection. However this method can only compare a given list of candidate models instead of searching in the exponentially large structure space.

The proposed MCMC method is shown to provide accurate posterior distributions at small step sizes. The selective shrinkage property of the spike and slab prior enables us to learn an MRF at different sparsity levels without noticeably suffering from under-fitting or over-fitting even for a small data set. Experiments with simulated data and real-world data show that the Bayesian method can learn both an accurate structure and a set of parameter values with strong predictive performance. In contrast the $L_1$-based methods could fail to accomplish both tasks with a single choice of the regularization strength. Also, the performance of our Bayesian model is largely insensitive to the choice of hyper-parameters. It provides an automated way to choose a proper sparsity level, while $L_1$ methods usually rely on cross-validation to find their optimal regularization setting.

\section*{Acknowledgement}
This material is based upon work supported by the National Science Foundation under Grant No. 0914783, 0928427, 1018433.

\newpage

\bibliography{Refs_yutian.bib}

\begin{thebibliography}{34}
\providecommand{\natexlab}[1]{#1}
\providecommand{\url}[1]{\texttt{#1}}
\expandafter\ifx\csname urlstyle\endcsname\relax
  \providecommand{\doi}[1]{doi: #1}\else
  \providecommand{\doi}{doi: \begingroup \urlstyle{rm}\Url}\fi

\bibitem[Andrew and Gao(2007)]{andrew2007scalable}
G.~Andrew and J.~Gao.
\newblock Scalable training of {L}1-regularized log-linear models.
\newblock In \emph{Proceedings of the 24th international conference on Machine
  learning}, pages 33--40. ACM, 2007.

\bibitem[Bresler et~al.(2008)Bresler, Mossel, and
  Sly]{bresler2008reconstruction}
G.~Bresler, E.~Mossel, and A.~Sly.
\newblock Reconstruction of {M}arkov random fields from samples: Some
  observations and algorithms.
\newblock \emph{Approximation, Randomization and Combinatorial Optimization.
  Algorithms and Techniques}, pages 343--356, 2008.

\bibitem[Courville et~al.(2011)Courville, Bergstra, and
  Bengio]{courville2011spike}
A.~Courville, J.~Bergstra, and Y.~Bengio.
\newblock A spike and slab restricted {B}oltzmann machine.
\newblock \emph{Journal of Machine Learning Research, W\&CP}, 15, 2011.

\bibitem[Della~Pietra et~al.(1997)Della~Pietra, Della~Pietra, and
  Lafferty]{della1997inducing}
S.~Della~Pietra, V.~Della~Pietra, and J.~Lafferty.
\newblock Inducing features of random fields.
\newblock \emph{Pattern Analysis and Machine Intelligence, IEEE Transactions
  on}, 19\penalty0 (4):\penalty0 380--393, 1997.

\bibitem[Dudik et~al.(2004)Dudik, Phillips, and Schapire]{dudik2004performance}
M.~Dudik, S.~Phillips, and R.~Schapire.
\newblock Performance guarantees for regularized maximum entropy density
  estimation.
\newblock \emph{Learning Theory}, pages 472--486, 2004.

\bibitem[Friedman et~al.(2010)Friedman, Hastie, and
  Tibshirani]{friedman2010regularization}
Jerome Friedman, Trevor Hastie, and Robert Tibshirani.
\newblock Regularization paths for generalized linear models via coordinate
  descent.
\newblock \emph{Journal of Statistical Software}, 33\penalty0 (1):\penalty0
  1--22, 2010.
\newblock URL \url{http://www.jstatsoft.org/v33/i01/}.

\bibitem[Girolami and Calderhead(2011)]{girolami2011riemann}
M.~Girolami and B.~Calderhead.
\newblock Riemann manifold {L}angevin and {H}amiltonian {M}onte {C}arlo
  methods.
\newblock \emph{Journal of the Royal Statistical Society: Series B (Statistical
  Methodology)}, 73\penalty0 (2):\penalty0 123--214, 2011.

\bibitem[Green(1995)]{green1995reversible}
P.J. Green.
\newblock Reversible jump {M}arkov chain {M}onte {C}arlo computation and
  {B}ayesian model determination.
\newblock \emph{Biometrika}, 82\penalty0 (4):\penalty0 711--732, 1995.

\bibitem[Hern{\'a}ndez-Lobato et~al.(2010)Hern{\'a}ndez-Lobato,
  Hern{\'a}ndez-Lobato, Helleputte, and Dupont]{hernández2010expectation}
D.~Hern{\'a}ndez-Lobato, J.~Hern{\'a}ndez-Lobato, T.~Helleputte, and P.~Dupont.
\newblock Expectation propagation for {B}ayesian multi-task feature selection.
\newblock \emph{Machine Learning and Knowledge Discovery in Databases}, pages
  522--537, 2010.

\bibitem[H{\"o}fling and Tibshirani(2009)]{höfling2009estimation}
H.~H{\"o}fling and R.~Tibshirani.
\newblock Estimation of sparse binary pairwise {M}arkov networks using
  pseudo-likelihoods.
\newblock \emph{The Journal of Machine Learning Research}, 10:\penalty0
  883--906, 2009.

\bibitem[Horowitz(1991)]{horowitz1991generalized}
A.M. Horowitz.
\newblock A generalized guided {M}onte {C}arlo algorithms.
\newblock \emph{Physics Letters B}, 268\penalty0 (2):\penalty0 247--252, 1991.

\bibitem[Hu et~al.(2009)Hu, Joshi, and Johnson]{hu2009log}
J.~Hu, A.~Joshi, and V.E. Johnson.
\newblock Log-linear models for gene association.
\newblock \emph{Journal of the American Statistical Association}, 104\penalty0
  (486):\penalty0 597--607, 2009.

\bibitem[Ishwaran and Rao(2005)]{ishwaran2005spike}
H.~Ishwaran and J.S. Rao.
\newblock Spike and slab variable selection: frequentist and {B}ayesian
  strategies.
\newblock \emph{The Annals of Statistics}, 33\penalty0 (2):\penalty0 730--773,
  2005.

\bibitem[Jones et~al.(2005)Jones, Carvalho, Dobra, Hans, Carter, and
  West]{jones2005experiments}
B.~Jones, C.~Carvalho, A.~Dobra, C.~Hans, C.~Carter, and M.~West.
\newblock Experiments in stochastic computation for high-dimensional graphical
  models.
\newblock \emph{Statistical Science}, 20\penalty0 (4):\penalty0 388--400, 2005.

\bibitem[Lee et~al.(2006)Lee, Ganapathi, and Koller]{lee2006efficient}
S.I. Lee, V.~Ganapathi, and D.~Koller.
\newblock Efficient structure learning of {M}arkov networks using {L}1
  regularization.
\newblock In \emph{In NIPS}. Citeseer, 2006.

\bibitem[Li(2009)]{li2009markov}
S.Z. Li.
\newblock \emph{Markov random field modeling in image analysis}.
\newblock Springer-Verlag New York Inc, 2009.

\bibitem[Lin and Lee(2006)]{lin2006bayesian}
Y.~Lin and D.D. Lee.
\newblock {B}ayesian {L}1-norm sparse learning.
\newblock In \emph{Acoustics, Speech and Signal Processing, 2006. ICASSP 2006
  Proceedings. 2006 IEEE International Conference on}, volume~5, pages V--V.
  IEEE, 2006.

\bibitem[Mitchell and Beauchamp(1988)]{mitchell1988bayesian}
T.J. Mitchell and J.J. Beauchamp.
\newblock Bayesian variable selection in linear regression.
\newblock \emph{Journal of the American Statistical Association}, pages
  1023--1032, 1988.

\bibitem[Mohamed et~al.(2011)Mohamed, Heller, and
  Ghahramani]{mohamed2011bayesian}
S.~Mohamed, K.~Heller, and Z.~Ghahramani.
\newblock {B}ayesian and {L}1 approaches to sparse unsupervised learning.
\newblock \emph{Arxiv preprint arXiv:1106.1157}, 2011.

\bibitem[Murray and Ghahramani(2004)]{MurrayGhahramani04}
I.~Murray and Z.~Ghahramani.
\newblock Bayesian learning in undirected graphical models: approximate {MCMC}
  algorithms.
\newblock In \emph{Proceedings of the 14th Annual Conference on Uncertainty in
  AI}, pages 392--399, 2004.

\bibitem[Murray et~al.(2006)Murray, Ghahramani, and
  MacKay]{MurrayGhahramaniMacKay06}
I.~Murray, Z.~Ghahramani, and D.J.C. MacKay.
\newblock {MCMC} for doubly-intractable distributions.
\newblock In \emph{Proceedings of the 14th Annual Conference on Uncertainty in
  Artificial Intelligence (UAI-06)}, Pittsburgh, PA, 2006.

\bibitem[Neal(2010)]{neal2010mcmc}
R.M. Neal.
\newblock {MCMC} using {H}amiltonian dynamics.
\newblock \emph{Handbook of Markov Chain Monte Carlo: Methods and
  Applications}, page 113, 2010.

\bibitem[Parise and Welling(2006)]{parise2006structure}
S.~Parise and M.~Welling.
\newblock Structure learning in {M}arkov random fields.
\newblock \emph{Advances in Neural Information Processing Systems}, 2006.

\bibitem[Park and Casella(2008)]{park2008bayesian}
T.~Park and G.~Casella.
\newblock The {B}ayesian lasso.
\newblock \emph{Journal of the American Statistical Association}, 103\penalty0
  (482):\penalty0 681--686, 2008.

\bibitem[Qi et~al.(2005)Qi, Szummer, and Minka]{QiSzummerMinka05}
Y.~Qi, M.~Szummer, and T.P. Minka.
\newblock Bayesian conditional random fields.
\newblock In \emph{Artificial Intelligence and Statistics}, 2005.

\bibitem[Ravikumar et~al.(2010)Ravikumar, Wainwright, and
  Lafferty]{ravikumar2010high}
P.~Ravikumar, M.J. Wainwright, and J.D. Lafferty.
\newblock High-dimensional ising model selection using {L}1-regularized
  logistic regression.
\newblock \emph{The Annals of Statistics}, 38\penalty0 (3):\penalty0
  1287--1319, 2010.

\bibitem[Riezler and Vasserman(2004)]{riezler2004incremental}
S.~Riezler and A.~Vasserman.
\newblock Incremental feature selection and {L}1 regularization for relaxed
  maximum-entropy modeling.
\newblock In \emph{Proceedings of EMNLP}, volume~4, 2004.

\bibitem[Robins et~al.(2007)Robins, Pattison, Kalish, and
  Lusher]{robins2007introduction}
G.~Robins, P.~Pattison, Y.~Kalish, and D.~Lusher.
\newblock An introduction to exponential random graph models for social
  networks.
\newblock \emph{Social networks}, 29\penalty0 (2):\penalty0 173--191, 2007.

\bibitem[Sha and Pereira(2003)]{sha2003shallow}
F.~Sha and F.~Pereira.
\newblock Shallow parsing with conditional random fields.
\newblock In \emph{Proceedings of the 2003 Conference of the North American
  Chapter of the Association for Computational Linguistics on Human Language
  Technology-Volume 1}, pages 134--141. Association for Computational
  Linguistics, 2003.

\bibitem[Tieleman(2008)]{Tieleman08}
T.~Tieleman.
\newblock Training restricted {B}oltzmann machines using approximations to the
  likelihood gradients.
\newblock In \emph{Proceedings of the International Conference on Machine
  Learning}, volume~25, pages 1064--1071, 2008.

\bibitem[Tsuruoka et~al.(2009)Tsuruoka, Tsujii, and
  Ananiadou]{tsuruoka2009stochastic}
Y.~Tsuruoka, J.~Tsujii, and S.~Ananiadou.
\newblock Stochastic gradient descent training for {L}1-regularized log-linear
  models with cumulative penalty.
\newblock In \emph{Proceedings of the Joint Conference of the 47th Annual
  Meeting of the ACL and the 4th International Joint Conference on Natural
  Language Processing of the AFNLP: Volume 1-Volume 1}, pages 477--485.
  Association for Computational Linguistics, 2009.

\bibitem[Wainwright et~al.(2007)Wainwright, Ravikumar, and
  Lafferty]{wainwright2007high}
M.J. Wainwright, P.~Ravikumar, and J.D. Lafferty.
\newblock High-dimensional graphical model selection using {L}1-regularized
  logistic regression.
\newblock \emph{Advances in neural information processing systems},
  19:\penalty0 1465, 2007.

\bibitem[Welling and Teh(2011)]{welling2011bayesian}
M.~Welling and Y.W. Teh.
\newblock Bayesian learning via stochastic gradient langevin dynamics.
\newblock 2011.

\bibitem[Zhu et~al.(2010)Zhu, Lao, and Xing]{zhu2010grafting}
J.~Zhu, N.~Lao, and E.P. Xing.
\newblock Grafting-light: fast, incremental feature selection and structure
  learning of {M}arkov random fields.
\newblock In \emph{Proceedings of the 16th ACM SIGKDD international conference
  on Knowledge discovery and data mining}, pages 303--312. ACM, 2010.

\end{thebibliography}
\bibliographystyle{plainnat}

\end{document}